\documentclass[pdflatex,sn-mathphys-num]{sn-jnl}

\usepackage{graphicx}%
\usepackage{multirow}%
\usepackage{amsmath,amssymb,amsfonts}%
\usepackage{amsthm}%
\usepackage{mathrsfs}%
\usepackage[title]{appendix}%
\usepackage{xcolor}%
\usepackage{textcomp}%
\usepackage{manyfoot}%
\usepackage{booktabs}%
\usepackage{algorithm}%
\usepackage{algorithmicx}%
\usepackage{algpseudocode}%
\usepackage{listings}%
\usepackage{hyperref} %
\usepackage[most]{tcolorbox}

\theoremstyle{thmstyleone}%
\theoremstyle{thmstyletwo}%

\theoremstyle{thmstylethree}%

\usepackage{spverbatim}
\usepackage{listings}
\usepackage{caption}

\definecolor{codegreen}{rgb}{0,0.6,0}
\definecolor{codegray}{rgb}{0.5,0.5,0.5}
\definecolor{codepurple}{rgb}{0.58,0,0.82}
\definecolor{backcolour}{rgb}{0.95,0.95,0.92}
\definecolor{lightgray}{gray}{0.8}

\lstdefinestyle{codestyle}{
	backgroundcolor=\color{backcolour},   
	commentstyle=\color{codegreen},
	keywordstyle=\color{magenta},
	numberstyle=\tiny\color{codegray},
	stringstyle=\color{codepurple},
	basicstyle=\ttfamily\tiny,
	breakatwhitespace=false,         
	breaklines=true,                 
	captionpos=b,                    
	keepspaces=true,                 
	numbers=left,                    
	showspaces=false,                
	showstringspaces=false,
	showtabs=false,                  
	tabsize=2,
	frame=single,
	rulecolor=\color{black!30},
	framesep=5pt,
	frameround=tttt,
	language=Python,
	numbersep=10pt,
	xleftmargin=10pt,                 
	xrightmargin=10pt,
	literate={\#}{{\textcolor{magenta}{\#}}}1
}
\newcounter{codecounter}
\lstnewenvironment{code}[2][]{
	\refstepcounter{codecounter}
	
	\let\savelstlisting\thelstlisting
	\renewcommand{\thelstlisting}{\arabic{codecounter}}
	\lstset{
		style=codestyle,
		caption={Code \thecodecounter: #1}, 
		label={#2},
		title={Code~\thecodecounter: #1}
	}
}{
	\let\thelstlisting\savelstlisting
}

\lstdefinestyle{promptstyle}{
	backgroundcolor=\color{yellow!5},      
	basicstyle=\ttfamily\tiny\raggedright,       
	breakatwhitespace=false,
	breaklines=true,
	captionpos=b,
	keepspaces=true,
	numbers=none,                       
	showspaces=false,
	showstringspaces=false,
	showtabs=false,
	tabsize=4,
	frame=single,
	rulecolor=\color{orange!40},   
	framesep=8pt,
	frameround=tttt,
	language=,                         
	numbersep=0pt,
	xleftmargin=10pt,                 
	xrightmargin=10pt,
	aboveskip=10pt,   
	belowskip=10pt,
	escapechar=@,
	columns=fullflexible,              
	upquote=true,                     
	literate={\#}{{\textcolor{orange!80}{\#}}}1,
	postbreak=\space
}

\newcounter{promptcounter}
\lstnewenvironment{prompt}[2][]{
	\refstepcounter{promptcounter}
	
	\let\savelstlisting\thelstlisting
	\renewcommand{\thelstlisting}{\arabic{promptcounter}}
	\lstset{
		style=promptstyle,
		caption={Prompt \thepromptcounter: #1}, 
		label={#2},
		title={Prompt~\thepromptcounter: #1}
	}
}{
	\let\thelstlisting\savelstlisting
}

\usepackage[T1]{fontenc}
\usepackage{graphicx}
\usepackage[absolute,overlay]{textpos}
\begin{document}
	
\begin{textblock}{}(15, 15)
	\includegraphics[width=3cm]{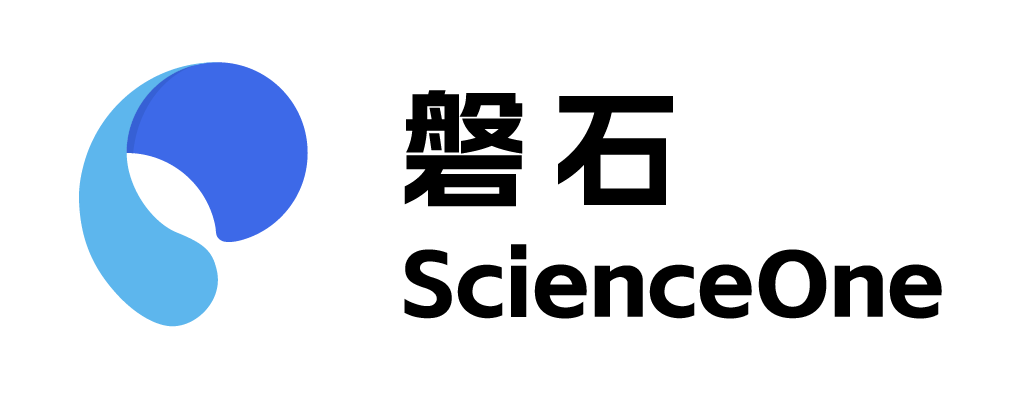}\includegraphics[width=3.5cm]{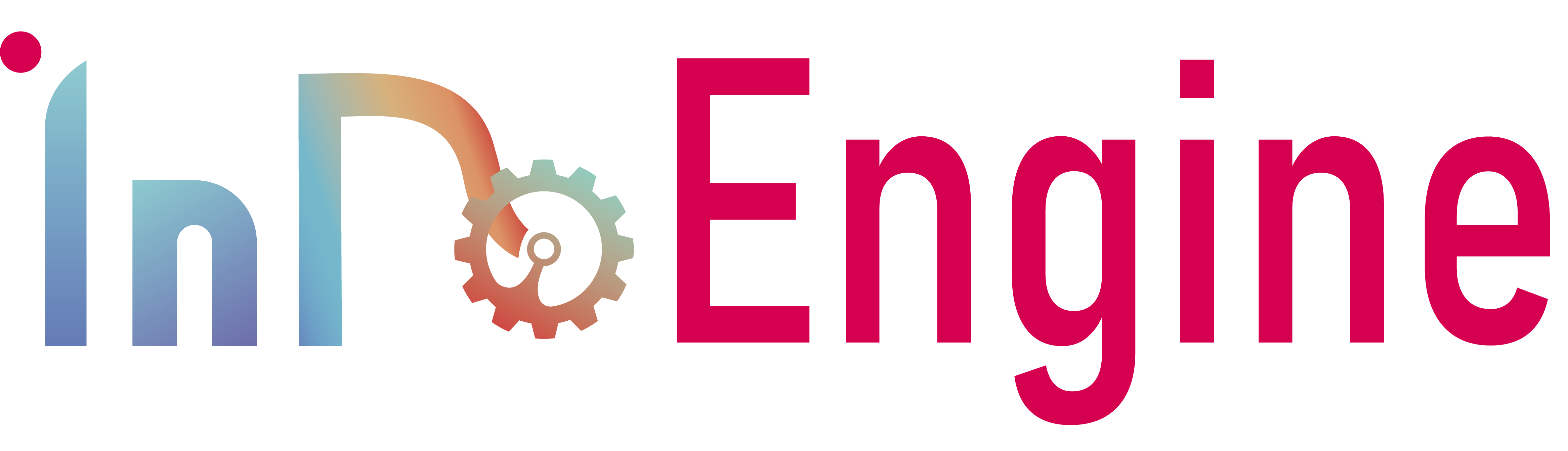}
	\vspace{1mm}
	\textcolor{gray!60}{\rule{0.9\paperwidth}{1pt}}
\end{textblock}

\title[InfEngine]{InfEngine: A Self-Verifying and Self-Optimizing Intelligent Engine for Infrared Radiation Computing}


\author[1]{\fnm{Kun} \sur{Ding}}\email{kun.ding@ia.ac.cn}
\author[1]{\fnm{Jian} \sur{Xu}}\email{jian.xu@ia.ac.cn}
\author*[1]{\fnm{Ying} \sur{Wang}}\email{ywang@nlpr.ia.ac.cn}
\author[1]{\fnm{Peipei} \sur{Yang}}\email{ppyang@nlpr.ia.ac.cn}
\author[1]{\fnm{Shiming} \sur{Xiang}}\email{smxiang@nlpr.ia.ac.cn}

\affil[1]{\orgdiv{State Key Laboratory of Multimodal Artificial Intelligence Systems}, \orgname{Institute of Automation, Chinese Academy of Sciences}, \orgaddress{\city{Beijing}, \country{China}}}


\abstract{Infrared radiation computing underpins advances in climate science, remote sensing and spectroscopy but remains constrained by manual workflows. We introduce \textbf{InfEngine}, an autonomous intelligent computational engine designed to drive a paradigm shift from human-led orchestration to collaborative automation. It integrates four specialized agents through two core innovations: \textbf{self-verification}, enabled by joint solver-evaluator debugging, improves functional correctness and scientific plausibility; \textbf{self-optimization}, realized via evolutionary algorithms with self-discovered fitness functions, facilitates autonomous performance optimization. Evaluated on InfBench with \textbf{200} infrared-specific tasks and powered by InfTools with \textbf{270} curated tools, InfEngine achieves a \textbf{92.7\%} pass rate and delivers workflows \textbf{21$\times$} faster than manual expert effort. More fundamentally, it illustrates how researchers can transition from manual coding to collaborating with self-verifying, self-optimizing computational partners. By generating reusable, verified and optimized code, InfEngine transforms computational workflows into persistent scientific assets, accelerating the cycle of scientific discovery.}




\maketitle

\section{Introduction}\label{sec:intro}
Infrared radiation constitutes a critical segment of the electromagnetic spectrum, from 780~$\mathrm{nm}$ to 1~$\mathrm{mm}$. The computational modeling, simulation and analysis of this radiation, termed infrared radiation computing, constitutes a foundational technical area that underpins progress across diverse scientific and engineering disciplines. It enables screening of novel photodetector materials \cite{yin2018ultrafast,cheng20222d} and supports elucidation of molecular structures from spectral fingerprints \cite{alberts2024leveraging,NMIRacle}. It also facilitates atmospheric remote sensing \cite{CLOUGH2005233} and drives development of advanced thermal imaging systems \cite{10014654}.

\begin{figure*}[!h]
	\centering
	\includegraphics[width=0.95\textwidth]{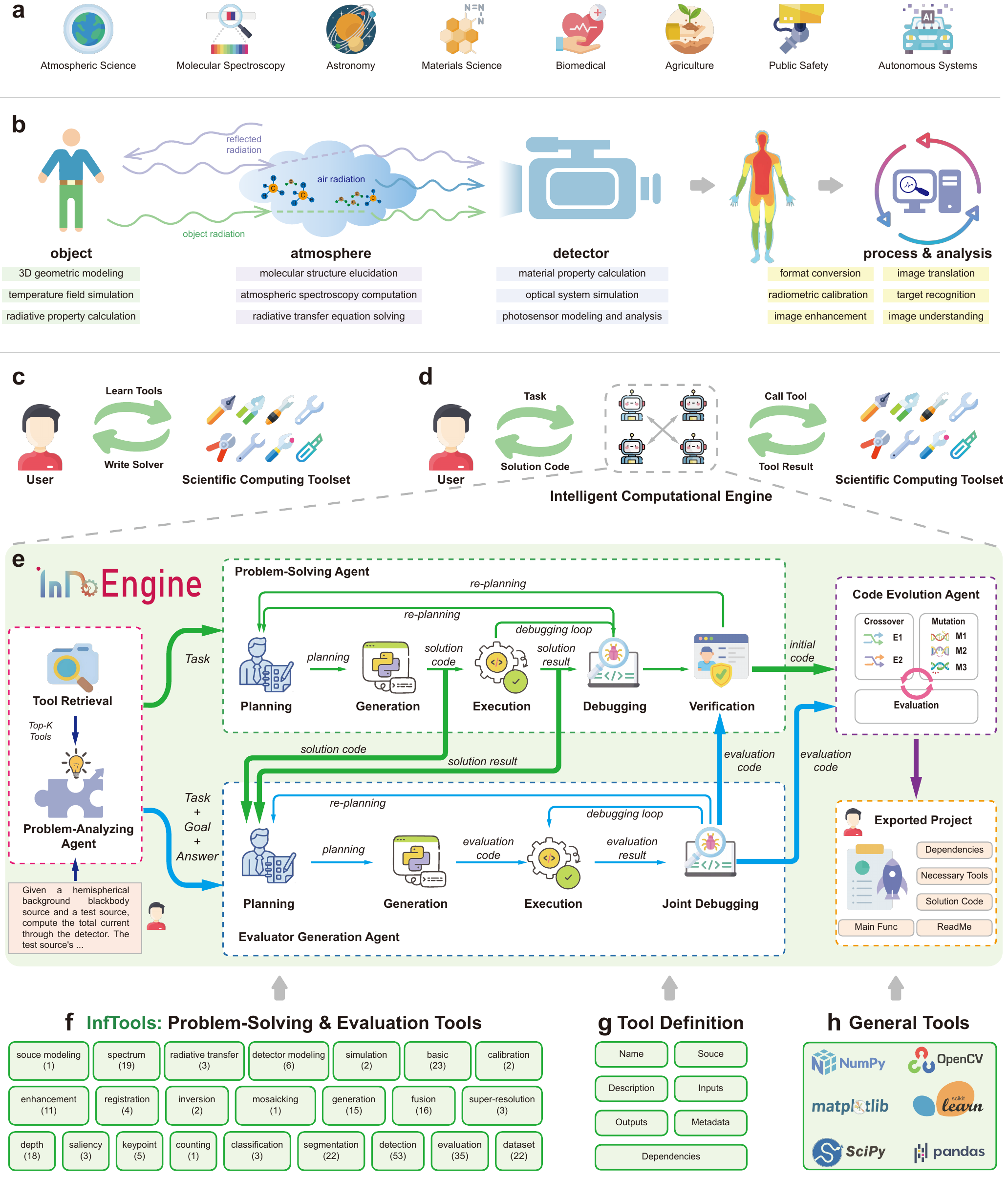}
	\caption{Overview. \textbf{a}, Related applications. \textbf{b}, Workflow of infrared radiation computing with core computational components at each stage. \textbf{c}, Scientific computing based on manual coding. \textbf{d}, Scientific computing based on intelligent computing engine. \textbf{e}, The flowchart of \textbf{InfEngine}. \textbf{f}, The \textbf{InfTools} has 270 tools for infrared radiation computing. \textbf{g}, Standardized fields defined for each tool. \textbf{h}, Some of the general tools that can be used in InfEngine. Icon credit: \href{https://www.freepik.com/}{Freepik.com}.}
	\label{fig:overview}
\end{figure*}

A complete infrared radiation computing workflow comprises a multi-stage pipeline (Fig.~\ref{fig:overview}\textcolor{blue}{b}), which starts with source modeling and spectroscopic analysis, proceeds through radiative transfer simulation in various media, and ends with data processing and physical interpretation. Currently, constructing such integrated workflows is predominantly a manual task. Researchers must act as the central ``orchestrator and validator'' (Fig.~\ref{fig:overview}\textcolor{blue}{c}) and manually integrate disparate software packages \cite{Leroy_Eradiate_radiative_transfer_2025,emde2016libradtran}, numerical solvers \cite{du2023fast} and physical models \cite{Marzano2014}. This labor-intensive process requires deep domain expertise for coding, integration and validation. It creates a significant bottleneck that impedes rapid iteration and limits scalability. Ultimately, it hinders the transformative potential of infrared radiation computing across its broad applications, highlighting an urgent need for intelligent automation.

Recent advances in large language models (LLMs) \cite{NEURIPS2020_1457c0d6} offer promising avenues for automation in related domains. Two dominant approaches have emerged. The first is \textbf{tool calling} \cite{QuDWCWYXW25,lu2025octotools}, which empowers LLMs to select and sequence tools from a predefined library. This paradigm has been successfully applied across multiple domains, including scientific reasoning (SciAgent \cite{SciAgent}, AWL \cite{lyu2025adapting}), chemistry (ChemCrow \cite{ChemCrow}, ChatMOF \cite{ChatMOF}), genomics (GeneGPT \cite{GeneGPT}) and differential equation solving (PDE-Agent \cite{PDE-Agent}). The second is \textbf{code generation}. It produces executable scripts and offers greater flexibility for composing complex operations. This paradigm has advanced general programming through methods like CodeCoT \cite{huang2024codecottacklingcodesyntax}. It has even facilitated automated algorithm discovery, as seen in works such as FunSearch \cite{FunSearch}, EoH \cite{EoH}, EvoVLMA~\cite{EvoVLMA} and Evo-MCTS~\cite{EvoMCTS}.

Despite their successes in other domains, both paradigms have significant limitations for infrared radiation computing. Tool-calling approaches rely on a fixed set of atomic operations. This restricts their ability to express intricate algorithmic logic, integrate domain-specific tools with general-purpose libraries seamlessly, and generate reusable software artifacts. Code-generation approaches are more expressive but often prioritize syntactic correctness over functional and scientific validity. They may produce error-free code that nonetheless yields physically implausible results---a perilous silent failure in computational science. Furthermore, they lack mechanisms for iterative self-optimization towards performance objectives.

To accelerate the paradigm shift in infrared radiation computing, we introduce \textbf{InfEngine}, an \textbf{intelligent computational engine }inherently designed for \textbf{self-verification} and \textbf{self-optimization}. Our work builds on two foundational resources. The first is \textbf{InfTools}, a curated suite of \textbf{270} domain-specific tools covering the full spectrum of infrared radiation computing tasks. The second is \textbf{InfBench}, a new benchmark of \textbf{200} tasks tailored to evaluate both \textit{assistant-type} and \textit{optimization-type} problem-solving capabilities. To realize its core functionalities, InfEngine adopts a multi-agent architecture consisting of four specialized agents: the Problem Analyzing Agent for formalizing user queries, the Problem Solving Agent for generating solution code, the Evaluator Generation Agent for defining validation criteria and the Code Evolution Agent for iteratively refining solutions. This well-structured design enables two key innovations.

The first is self-verification, achieved through a joint debugging process. The system generates both a solver and a corresponding evaluator. A dedicated referee diagnoses discrepancies and prevents spurious co-optimization, ensuring fixes maintain generalizability and fidelity to the true task constraints. This elevates validation from mere executability to semantically meaningful correctness assessment. The second is self-optimization, bootstrapped from this verified foundation. For optimization-type tasks, the validated evaluator is repurposed as a trustworthy fitness function within an evolutionary algorithm. Unlike prior work such as FunSearch~\cite{FunSearch}, which requires manually-specified objectives, our system autonomously synthesizes optimization goals from problem descriptions. This enables truly autonomous performance improvement.

Evaluated on InfBench, InfEngine demonstrates a practical step toward a new collaborative paradigm for scientific computing, where researchers interact not with passive toolkits but with autonomous, self-optimizing partners. This paradigm provides a blueprint for intelligent agents capable of accelerating discovery across diverse computational domains.

\section{Results}\label{sec:results}
\subsection{Experimental Setup}
As shown in Fig.~\ref{fig:overview}\textcolor{blue}{e}, InfEngine takes a text-described problem and generates code via multi-agent interactions. We compare against state-of-the-art baselines on InfBench. Since existing methods lack native tool support and are optimized for programming/math tasks, we augmented their prompts with retrieved tool descriptions and output formats, leaving their core algorithms unchanged. For InfEngine, we set evolutionary iterations $N=10$ and population size $m=5$. We also evaluate across different LLMs.

\subsection{Performance on InfBench}\label{subsec2}
\subsubsection{Comparison with Different Methods}
We benchmark InfEngine against established code generation methods: Direct that prompts LLM directly, FewShot that augments Direct with two examples, CodeCoT~\cite{huang2024codecottacklingcodesyntax}, SelfDebug~\cite{ChenLSZ24}, Reflexion~\cite{ShinnCGNY23} and MapCoder~\cite{IslamAP24}. We adopt retrieval-augmented generation (RAG) to supply each approach with the top-15 most relevant tools. All experiments are conducted on Qwen3-8B, deployed via vLLM.

\begin{figure*}[!th]
	\centering
	\includegraphics[width=0.95\textwidth]{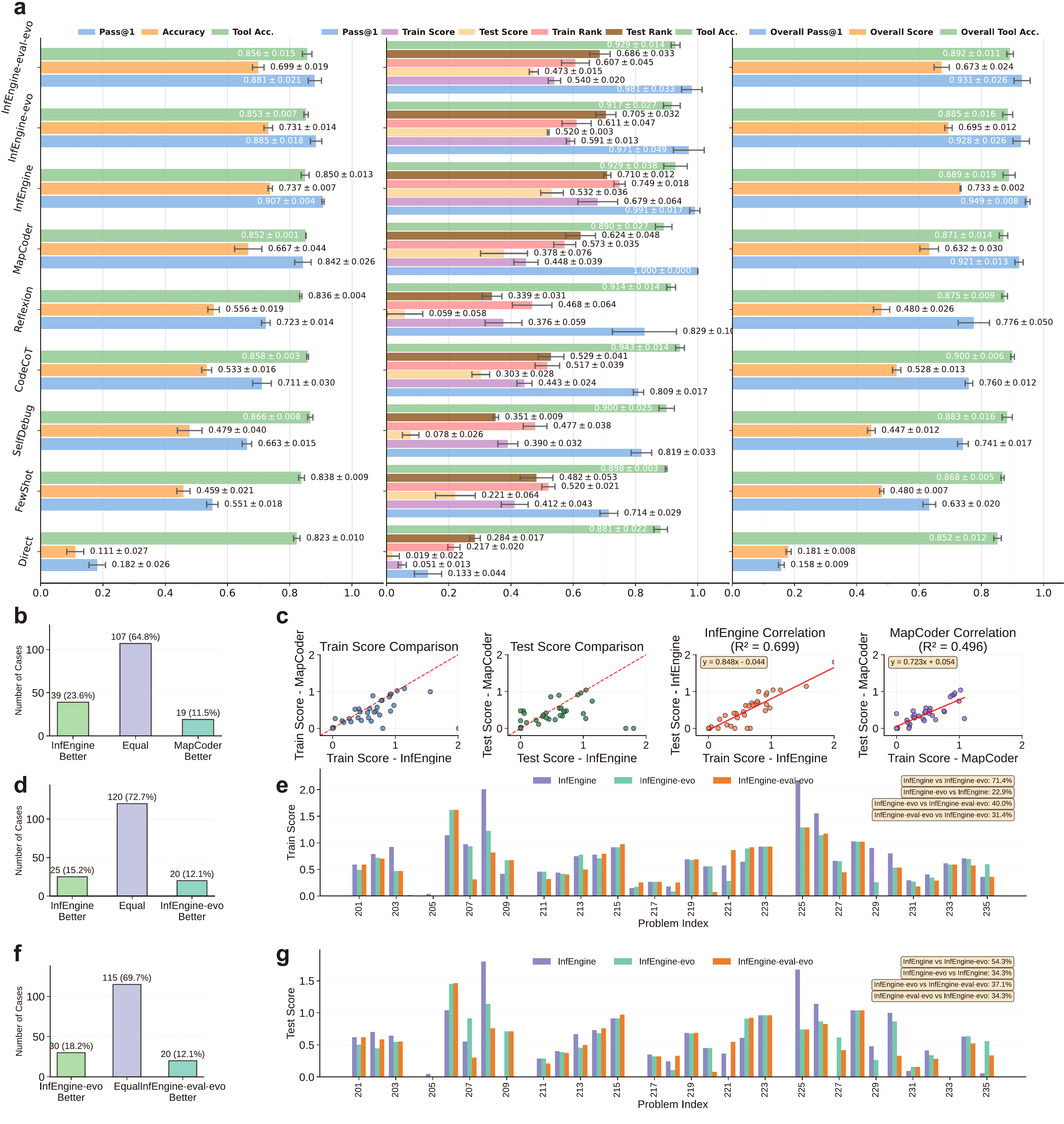}
	\caption{Comparison with baselines. \textbf{a}, Performance on assistant-type tasks (left), optimization-type tasks (mid), overall performance (right). \textbf{b}, Win rate comparison of InfEngine and MapCoder on assistant-type tasks. \textbf{c}, Train Score scatter plot of InfEngine and MapCoder on optimization-type tasks, Test Score scatter plot of  them on optimization-type tasks, correlation analysis  between training and test scores of InfEngine, correlation analysis  between training and test scores of MapCoder. \textbf{d}, Win rate comparison of InfEngine and InfEngine-evo on assistant-type tasks. \textbf{e}, Per-task Train Score comparison of InfEngine, InfEngine-evo and InfEngine-eval-evo. \textbf{f}, Win rate comparison of InfEngine-evo and InfEngine-eval-evo on assistant-type tasks. \textbf{g}, Per-task Test Score comparison of InfEngine, InfEngine-evo and InfEngine-eval-evo.}
	\label{fig:diff_methods_compare}
\end{figure*}

Fig.~\ref{fig:diff_methods_compare} shows the results, where Direct and FewShot establish baseline performance, but are outperformed by more advanced methods. SelfDebug and Reflexion outperform these baselines, with particular effectiveness on assistant-type tasks. However, these two methods struggle with optimization scenarios and exhibit significant overfitting behavior. CodeCoT achieves a strong Overall Score of 0.528. Its balanced scores (Train Score=0.443, Test Score=0.303) indicate consistent performance across tasks. MapCoder achieves the strongest performance among baseline methods with an Overall Pass@1 of 0.921. Its normalized ranking scores (Train Rank=0.573, Test Rank=0.624) demonstrate strong performance relative to other baselines, supported by its raw scores (Train Score=0.448, Test Score=0.378). Its superior performance in assistant-type tasks (Pass@1=0.842) further confirms the value of emulating human development processes.

Our proposed InfEngine establishes a new state-of-the-art across all evaluation dimensions. As shown in Fig.~\ref{fig:diff_methods_compare}\textcolor{blue}{a}, it achieves a remarkable Overall Pass@1 of 0.949 and an Overall Score of 0.733, representing an absolute improvement of 2.8\% and a relative improvement of 16.0\% over MapCoder, respectively. This superiority is particularly pronounced in the more challenging optimization-type tasks, where InfEngine attains near-perfect Pass@1 (0.991) while simultaneously achieving the best training (Train Rank=0.749) and test (Test Rank=0.710) performance---an exceptional combination indicating both effectiveness and generalization capability. This is further evidenced by its strong raw performance (Train Score=0.679, Test Score=0.532). Notably, InfEngine maintains strong performance on assistant-type tasks with a Pass@1 of 0.907, surpassing MapCoder by 6.5\%.

For assistant-type tasks, the win-rate analysis (Fig.~\ref{fig:diff_methods_compare}\textcolor{blue}{b}) reveals that InfEngine matches or outperforms MapCoder in the vast majority (88.4\%) of tasks, with a clear win rate of 23.6\% against a loss rate of only 11.5\%. For optimization-type tasks, the per-task superiority is even more pronounced. The pairwise comparisons (Fig.~\ref{fig:diff_methods_compare}\textcolor{blue}{c}) show that the majority of data points fall below the diagonal line, indicating that InfEngine achieves higher training and test scores than MapCoder on most tasks. The consistent distribution of points below the diagonal in the test score comparison (Fig.~\ref{fig:diff_methods_compare}\textcolor{blue}{c}) provides direct visual evidence of InfEngine’s superior generalization at the task level.

A critical comparison of generalization behavior is shown in Fig.~\ref{fig:diff_methods_compare}\textcolor{blue}{c}. InfEngine exhibits a strong positive correlation between training and test scores ($R^2 = 0.699$), indicating that optimization success reliably predicts generalization. MapCoder, by contrast, shows markedly weaker correlation ($R^2 = 0.496$), reflecting poorer transfer and higher overfitting risk. These results demonstrate that InfEngine’s evolution with self-verification inherently yields more robust and generalizable solutions.

In summary, InfEngine's consistent superiority across both task types demonstrates its multi-agent evolutionary architecture fundamentally enhances the code generation process, producing robust and generalizable solutions that outperform existing methods.

\begin{figure*}[!t]
	\centering
	\includegraphics[width=0.95\textwidth]{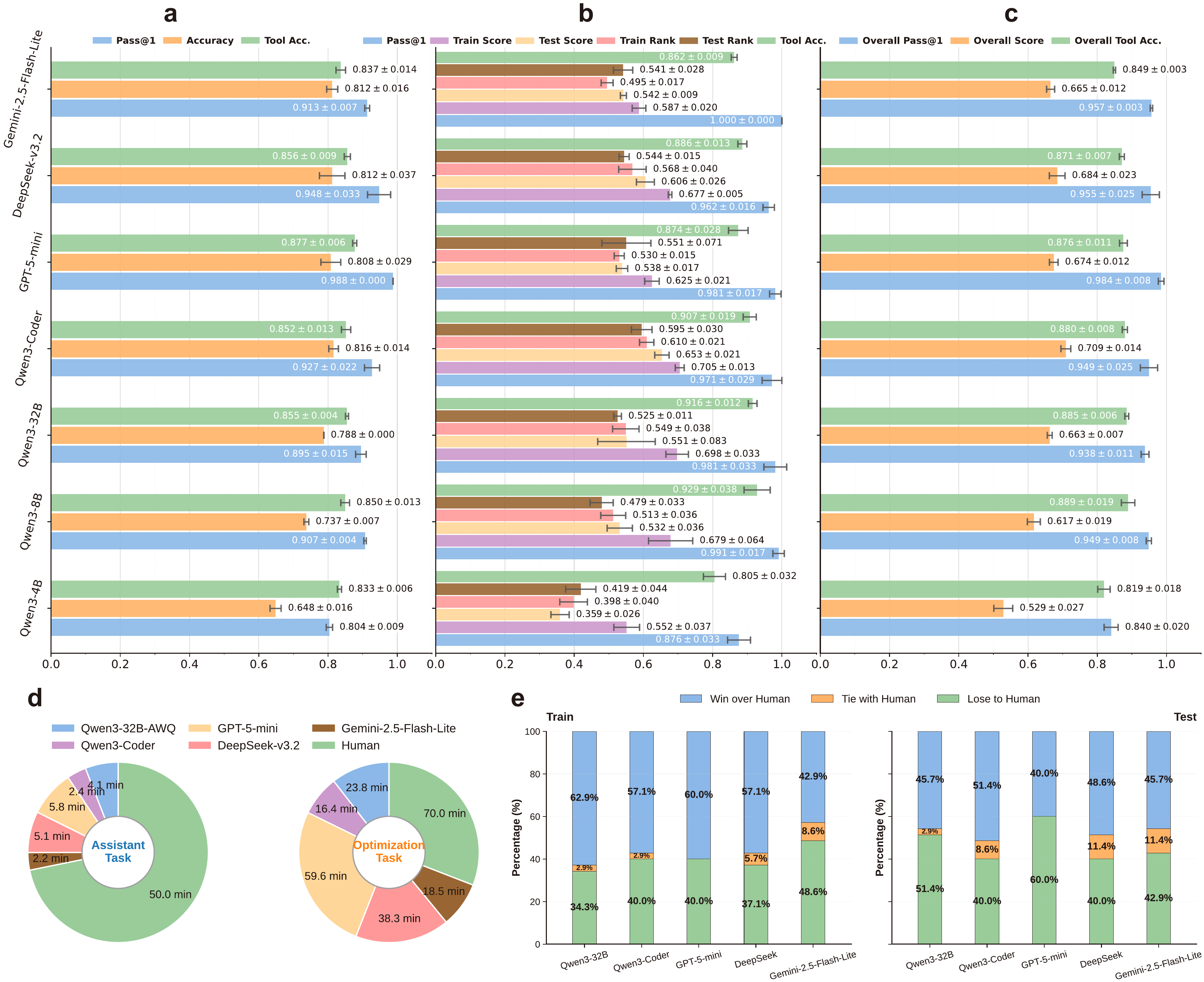}
	\caption{Comparison of InfEngine with different LLMs and human baseline. \textbf{a}, Performance on assistant-type tasks. \textbf{b}, Performance on optimization-type tasks. \textbf{c}, Aggregate performance across all tasks. \textbf{d}, Computational time comparison for assistant-type (left) and optimization-type (right) tasks. \textbf{e}, Win/Tie/Loss rate of different LLM-based InfEngine instances against the human baseline on the training set (left) and test set (right) for optimization-type tasks.}
	\label{fig:diff_LLMs_compare}
\end{figure*}

\subsubsection{Comparison with Different LLMs}
To evaluate InfEngine’s generality, we test it across seven LLMs of varying scale and specialization. Results in Fig.~\ref{fig:diff_LLMs_compare} reveal two key findings. \emph{First}, InfEngine delivers a strong performance floor even with small models. Qwen3-4B achieves an Overall Pass@1 of 0.840, a result competitive with or surpassing MapCoder on Qwen3-8B. This shows our framework better leverages base model capacity. \emph{Second}, performance scales clearly with model capability. Test Rank improves from 0.419 (Qwen3-4B) to 0.595 (Qwen3-Coder), and Train Rank from 0.398 to 0.610. InfEngine thus exploits stronger reasoning and generalization in larger models via its evolutionary loop.

\subsubsection{Comparison with Human}
We benchmark InfEngine against a human expert following a conventional manual coding workflow (Fig.~\ref{fig:overview}\textcolor{blue}{c}). Efficiency is measured by average task completion time and solution quality by evaluation scores. InfEngine delivers substantial efficiency gains (Fig.~\ref{fig:diff_LLMs_compare}\textcolor{blue}{d}). For assistant-type tasks, human experts required approximately 50 minutes per task, whereas InfEngine completed these tasks in 2.2 to 5.8 minutes, achieving 8.6-22.7$\times$ acceleration. For optimization-type tasks, human researchers averaged 70 minutes per task, while InfEngine finished in 16 to 60 minutes, achieving 1.2-4.4$\times$ speedup. This acceleration enables rapid hypothesis testing and parameter exploration at scales previously infeasible due to manual overhead.

Beyond efficiency, InfEngine demonstrates strong capability in autonomously discovering objectives and iteratively improving solution quality. Fig.~\ref{fig:diff_LLMs_compare}\textcolor{blue}{e} compares agent and human performance on optimization-type tasks. The combined win and tie rate indicates solutions at least as effective as human counterparts. On training set, this rate exceeded 50\% for all models, peaking at 65.7\% with Qwen3-32B. On test set, models including Qwen3-Coder-480B-A35B-Inst, DeepSeek-v3.2 and Gemini-2.5-Flash-Lite also surpassed 50\%, demonstrating robust generalization and performance consistently matching or exceeding human levels. These results confirm that InfEngine-generated solutions maintain, and often surpass, the quality of manually crafted code.

\subsubsection{Effectiveness of Components}
To evaluate core component contributions of InfEngine, we conduct an ablation study with two variants: (1) \texttt{InfEngine-evo} removes Code Evolution, retaining self-verification; (2) \texttt{InfEngine-eval-evo} further removes Evaluator Generation, reducing to baseline.

The overall performance profile in Fig.~\ref{fig:diff_methods_compare}\textcolor{blue}{a} establishes a clear performance gradient. On optimization-type tasks, performance follows: InfEngine $>$ InfEngine-evo $>$ InfEngine-eval-evo, demonstrating that each autonomous capability cumulatively enhances system performance in scenarios requiring refinement and objective pursuit. On assistant-type tasks, the ordering is InfEngine $\approx$ InfEngine-evo $>$ InfEngine-eval-evo. While both InfEngine and InfEngine-evo achieve comparable results for assistant-type tasks, the consistent superiority of the full system in optimization scenarios and overall metrics underscores the synergistic value of integrating both self-verification and self-optimization.

The value of self-verification is evident in Fig.~\ref{fig:diff_methods_compare}\textcolor{blue}{e-g}. On optimization-type tasks, verification alone (InfEngine-evo) confers a clear advantage over InfEngine-eval-evo: it achieves higher training scores in 40.0\% of tasks versus 31.4\% losses (Fig.~\ref{fig:diff_methods_compare}\textcolor{blue}{e}) and higher test scores in 37.1\% versus 34.3\% losses (Fig.~\ref{fig:diff_methods_compare}\textcolor{blue}{g}). This advantage also holds for assistant-type tasks (18.2\% wins vs. 12.1\% losses; Fig.~\ref{fig:diff_methods_compare}\textcolor{blue}{f}). These results confirm that the Evaluator Generation Agent establishes a reliable correctness criterion, yielding more accurate solutions even without iterative refinement.

Self-optimization contributes most decisively in optimization-type tasks (Fig.~\ref{fig:diff_methods_compare}\textcolor{blue}{e,g}). InfEngine substantially outperforms the verification-only variant (InfEngine-evo), achieving 71.4\% wins versus 22.9\% losses on training scores and 54.3\% wins versus 34.3\% losses on test scores. This demonstrates that the Code Evolution Agent, bootstrapped with a verified fitness function, effectively explores the solution space and iteratively improves performance while preserving generalization. On assistant-type tasks, the marginal win differential (15.2\% vs. 12.1\%; Fig.~\ref{fig:diff_methods_compare}\textcolor{blue}{d}) reflects that such tasks demand correct execution rather than quantitative optimization.

In summary, self-verification ensures correctness and self-optimization drives improvement, transforming InfEngine from a static generator into an adaptive engine that autonomously pursues and verifies higher-quality outcomes.

\subsection{Case Analysis}\label{subsec3}
\subsubsection{Case 1: High-Throughput Screening of Infrared Photodetector Materials}
The development of advanced infrared detectors demands efficient screening of semiconductors across multidimensional parameter spaces~\cite{yin2018ultrafast,cheng20222d}. Traditional approaches require expertise in radiative transfer, solid-state physics and numerical methods, creating steep barriers to rapid material assessment. This case illustrates how InfEngine accelerates detector performance evaluation through autonomous workflow generation.

\begin{figure*}[!t]
	\centering
	\includegraphics[width=0.95\textwidth]{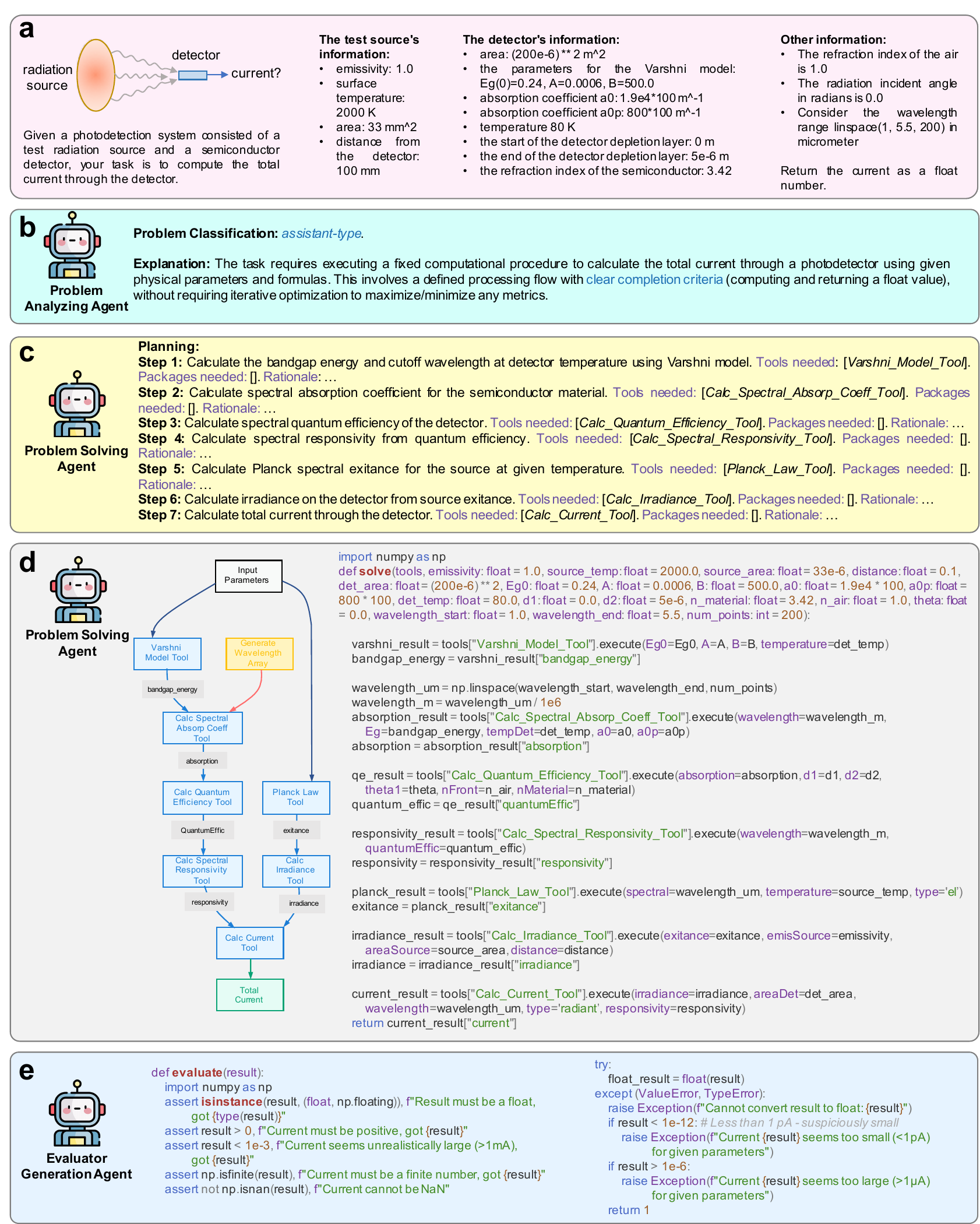}
	\caption{Question and result of Case 1. \textbf{a}, The query question with an illustration explaining the task. \textbf{b}, The result of Problem Analyzing Agent. \textbf{c}, The planning generated by Problem Solving Agent. \textbf{d}, The solution code and the corresponding flowchart.  \textbf{e}, Evaluation function generated by Evaluator Generation Agent. Credit: agent icon, \href{https://www.freepik.com/}{Freepik.com}.}
	\label{fig:case1}
\end{figure*}

Faced with a complex photocurrent calculation problem (Fig.~\ref{fig:case1}\textcolor{blue}{a}), InfEngine’s Problem Analyzing Agent first classified the task as deterministic computational pipeline rather than iterative optimization. The Problem Solving Agent then formulated a seven-step workflow (Fig.~\ref{fig:case1}\textcolor{blue}{c}), sequencing temperature-dependent bandgap modelling to photocurrent integration. The explicit stepwise reasoning reflects the system’s grasp of causal dependencies in optoelectronic systems. The generated Python code (Fig.~\ref{fig:case1}\textcolor{blue}{d}) faithfully implements this plan, handling unit conversion, spectral integration and interface effects. Its parameterized design with defaults matching the test case enables high-throughput screening: material parameters, device geometry and illumination conditions can be systematically varied while maintaining physical consistency. Crucially, InfEngine also autonomously produces a validation suite (Fig.~\ref{fig:case1}\textcolor{blue}{e}). The Evaluator Generation Agent embeds scientific constraints including current positivity, realistic amplitude spanning picoampere to microampere ranges, and numerical robustness directly into the evaluation pipeline, enhancing reliability in automated scientific computing.

By packaging complex optoelectronic physics into a parameterized, self-validating function, InfEngine transforms expert-led detector analysis into an automated and reproducible workflow. This capability accelerates the exploration of candidate materials across broad design spaces, advancing the discovery cycle for next-generation infrared sensing technologies.

\subsubsection{Case 2: Autonomous Cross-Validation of Computational Infrared Spectroscopy Methods}
Computational infrared spectroscopy relies on diverse methods~\cite{stienstra2024graphormer,bhatia2025leveraging} with distinct approximations and trade-offs. Directly comparing spectra predicted by different approaches, such as physics based molecular dynamics versus data driven structural inference, offers essential qualitative validation and reveals systematic biases. However, manually coordinating such comparative workflows across multiple simulation packages presents a substantial barrier. This case demonstrates how InfEngine automates this process by generating integrated pipelines that highlight spectral consistency and divergence across techniques.

\begin{figure*}[!t]
	\centering
	\includegraphics[width=\textwidth]{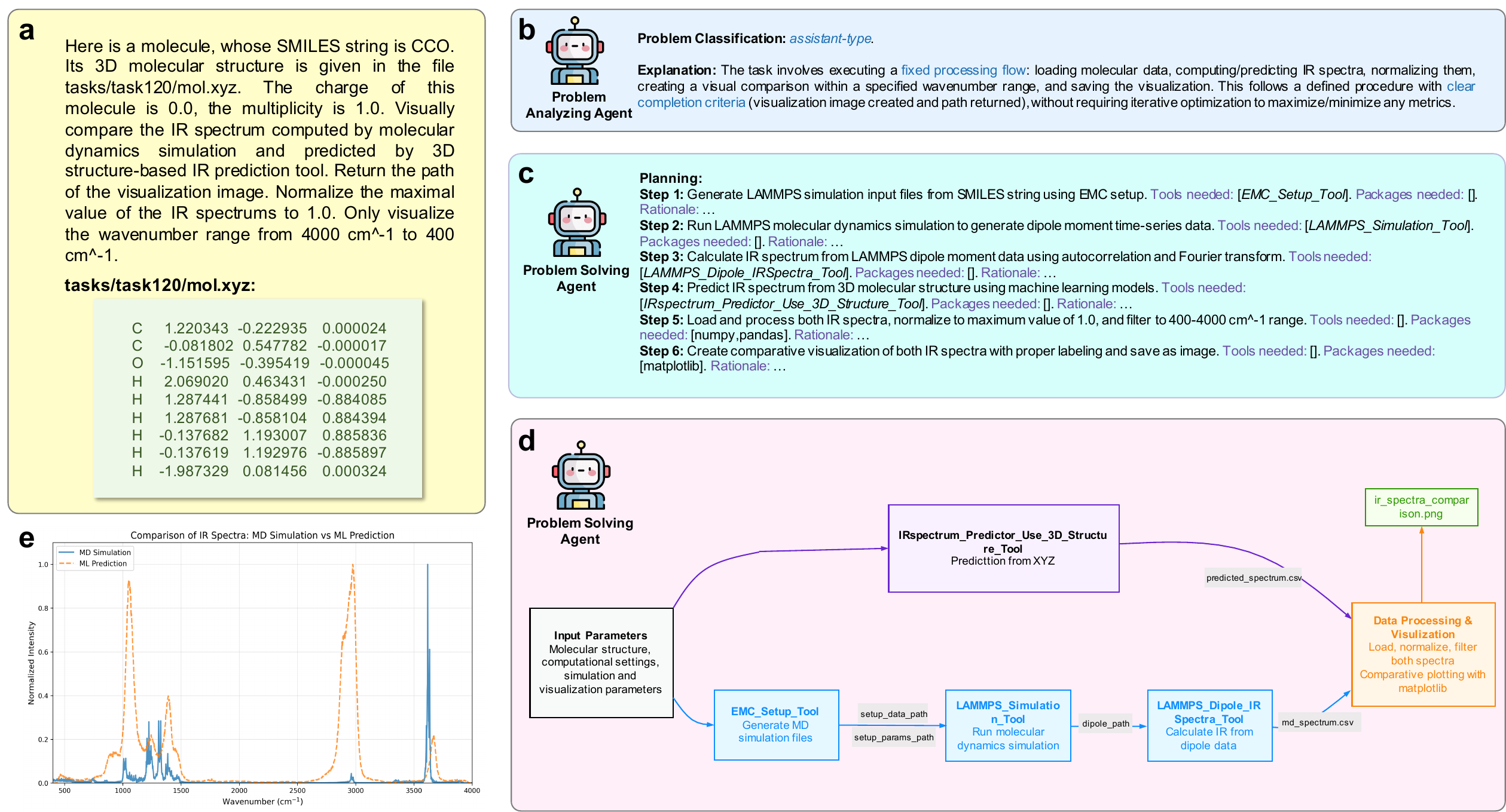}
	\caption{Question and result of Case 2. \textbf{a}, The query question requesting comparison of IR spectra from MD simulation and 3D structure-based prediction for ethanol (CCO). \textbf{b}, The result of Problem Analyzing Agent. \textbf{c}, The planning generated by Problem Solving Agent. \textbf{d}, The automated tool-chain and code execution flowchart of final solution code. \textbf{e}, The generated comparative visualization of normalized IR spectra by the solution code.}
	\label{fig:case2}
\end{figure*}

The task, as specified in Fig.~\ref{fig:case2}\textcolor{blue}{a}, involves a qualitative comparison of IR spectra for ethanol generated via two pathways: all atom molecular dynamics with LAMMPS and a machine learning prediction from the 3D molecular structure alone. The goal is not experimental benchmarking but automated workflow execution that visualizes and contrasts spectral features from both methods.

InfEngine constructed a workflow integrating four computational tools, autonomously linking a structure based IR predictor with a three stage MD simulation pipeline covering force field parameterization, trajectory generation and spectral postprocessing (Fig.~\ref{fig:case2}\textcolor{blue}{d}). This orchestration reflects the system's ability to translate a high-level comparative objective into a sequence of domain specific operations bridging distinct computational paradigms.

The generated code applies preprocessing steps including spectral filtering and intensity normalization to enable direct qualitative comparison. The resulting visualization (Fig.~\ref{fig:case2}\textcolor{blue}{e}) juxtaposes the two spectral traces in a publication ready format, allowing immediate assessment of peak positions, relative intensities and bandwidths as indicators of methodological agreement on the molecular vibrational fingerprint.

This case illustrates InfEngine's capacity to facilitate cross-validation of computational methods. By automating the setup, execution and analysis required for comparative spectroscopy, the system transforms a labor intensive process into a reproducible one click operation. The workflow delivers an immediate visual assessment of method consistency, serving as a vital sanity check and making qualitative validation routine and accessible in computational chemistry.

\begin{figure*}[!t]
	\centering
	\includegraphics[width=\textwidth]{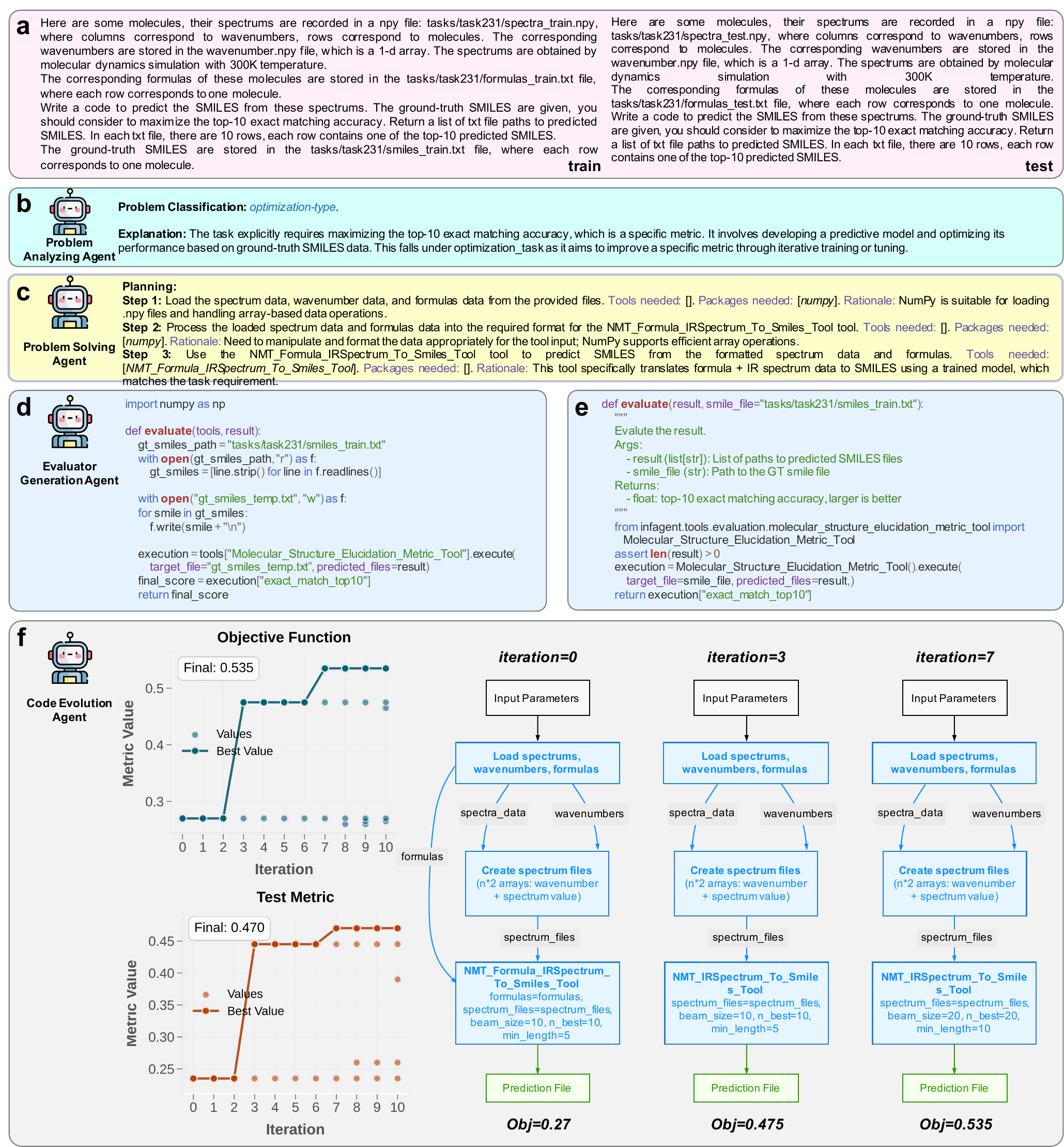}
	\caption{Question and result of Case 3. \textbf{a}, The query question (train+test). \textbf{b}, The output of Problem Analyzing Agent. \textbf{c}, The planning generated by Problem Solving Agent. \textbf{d}, The evaluation function generated by Evaluator Generation Agent. \textbf{e}, The evaluation code defined by human expert. \textbf{f}, Curves of generated evaluation metric on train set (as well as human evaluation metric on train set) and human evaluation metric on test set. Corresponding flowcharts of the solution code at iteration 0, 3,  7 are also included.}
	\label{fig:case3}
\end{figure*}

\subsubsection{Case 3: Autonomous Optimization of Molecular Structure Elucidation from Infrared Spectra}
Infrared spectra encode molecular vibrational fingerprints, yet reconstructing molecular structure from spectral data remains a long-standing inverse problem~\cite{MACE4IR,NMIRacle,alberts2024leveraging}. Traditional workflows require manual orchestration of preprocessing, feature extraction, and chemical space exploration through trial and error. This case demonstrates how InfEngine automates this optimization via autonomous workflow evolution.

Presented with the task of predicting SMILES from simulated IR spectra under a top-10 accuracy metric (Fig.~\ref{fig:case3}\textcolor{blue}{a}), InfEngine's Problem Analyzing Agent correctly identified an optimization-type task (Fig.~\ref{fig:case3}\textcolor{blue}{b}). The system generated a structured plan outlining data preparation and model prediction steps (Fig.~\ref{fig:case3}\textcolor{blue}{c}), and the Evaluator Generation Agent autonomously produced a tailored evaluation function to compute exact top-10 accuracy (Fig.~\ref{fig:case3}\textcolor{blue}{d}).

The initial solution employed a formula-integrated neural machine translation tool that required both spectra and molecular formulas as inputs, achieving an objective value of 0.270. Guided by iterative feedback, the Code Evolution Agent systematically refined the pipeline. By Iteration 3, the system autonomously discovered that molecular formulas provided limited utility and transitioned to a spectrum-only architecture, substantially improving performance to 0.475. Further optimization by Iteration 7 increased the beam size from 10 to 20 and minimum translation length from 5 to 10 to encourage more complete predictions, elevating the objective value to 0.535. This represents a relative improvement of 98\% from the initial solution (Fig.~\ref{fig:case3}\textcolor{blue}{f}).

This case illustrates InfEngine's capacity for autonomous methodological discovery. Through systematic search guided by objective evaluation, it transforms static spectroscopy workflows into adaptive, problem-specific solutions, identifying when to discard input modalities and how to adjust decoding parameters for optimal performance. This enables rapid domain adaptation without manual intervention and accelerates method development in molecular structure elucidation.

\begin{figure*}[!t]
	\centering
	\includegraphics[width=0.95\textwidth]{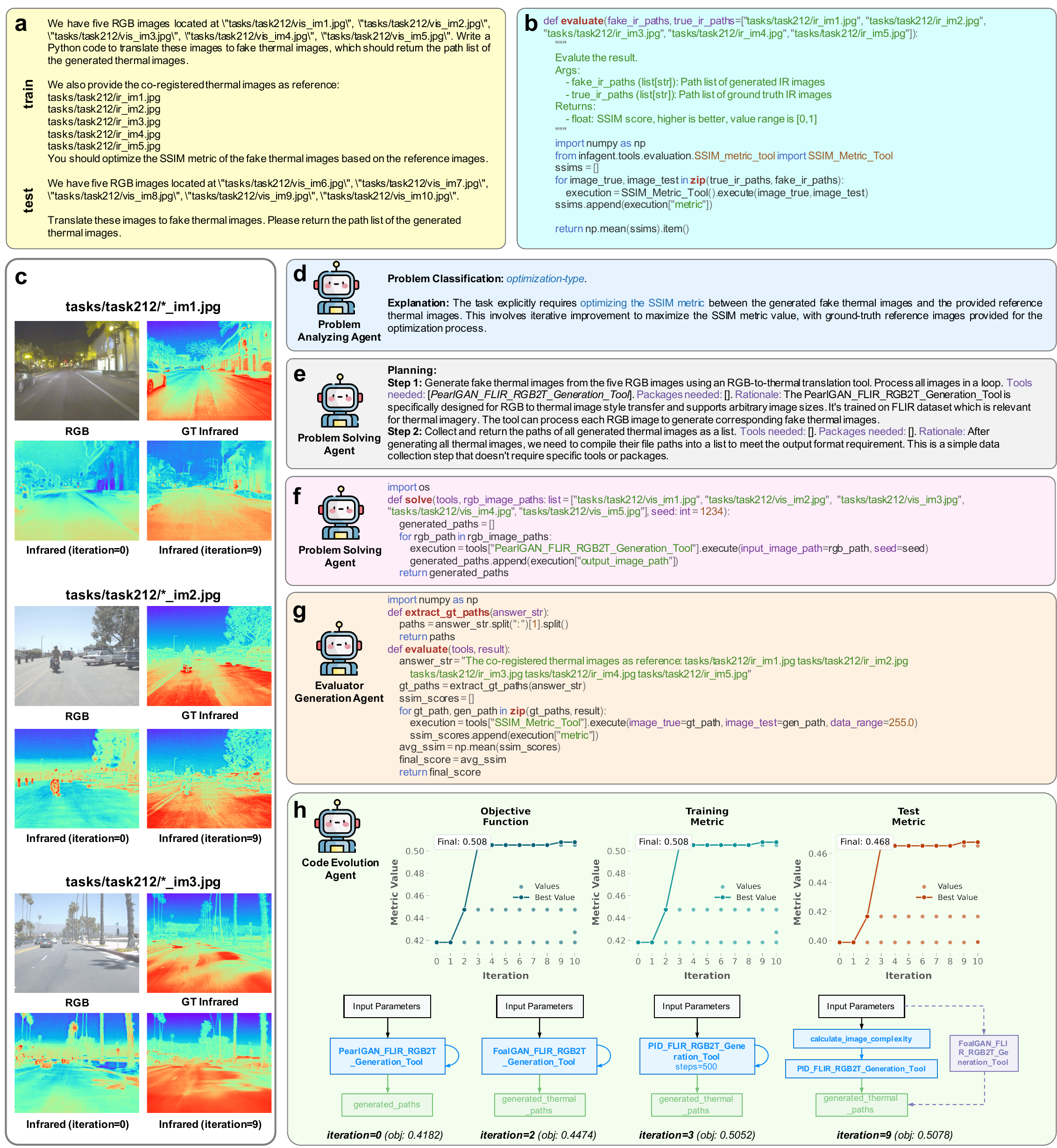}
	\caption{Question and result of case 4. \textbf{a}, The query question (train+test). \textbf{b}, The evaluation code written by human expert. \textbf{c}, Comparison of generated fake infrared images by solution code at iteration 0 and 9. \textbf{d}, The output of Problem Analyzing Agent. \textbf{e}, Planning generated by the Problem Solving Agent. \textbf{f}, The solution code generated by Problem Solving Agent. \textbf{g}, The generated evaluation code by Code Evolution Agent. \textbf{h}, Curves of generated evaluation metric on train set, human evaluation metric on train set and human evaluation metric on test set. Corresponding flowcharts of the solution code across different iterations are also included.}
	\label{fig:case4}
\end{figure*}

\subsection{Case 4: Evolutionary Optimization of Image Translation Pipeline}
Developing robust image-to-image translation systems for specialized domains like visible-to-thermal conversion requires navigating a vast space of algorithms and hyperparameters. Conventional approaches use fixed model selection or manual tuning, limiting adaptability. This case demonstrates how InfEngine autonomously discovers, evaluates and synthesizes an optimal translation pipeline through evolutionary code optimization, transitioning from tool invocation to adaptive system design.

As presented in Fig.~\ref{fig:case4}\textcolor{blue}{a}, the task required translating five RGB images into realistic thermal representations while maximizing SSIM against co-registered references. The Problem Solving Agent generated an initial plan (Fig.~\ref{fig:case4}\textcolor{blue}{e}) that established a scaffold for evolutionary refinement. The initial solution (Fig.~\ref{fig:case4}\textcolor{blue}{f}) employed PearlGAN\_FLIR\_RGB2T\_Generation\_Tool uniformly across images, achieving moderate SSIM (0.4182). Concurrently, the Evaluator Generation Agent autonomously produced code to compute SSIM (Fig.~\ref{fig:case4}\textcolor{blue}{g}), enabling quantitative feedback for optimization. Evolutionary search yielded progressive gains: replacing PearlGAN with FoalGAN raised SSIM to 0.4474, and adopting PID\_FLIR\_RGB2T\_Generation\_Tool with steps=500 further improved it to 0.5052. By iteration 9, the system synthesized an adaptive pipeline achieving 0.5078 SSIM. It introduced a content-aware module estimating edge density to dynamically adjust diffusion steps, and a fallback mechanism ensuring robustness when preferred tools failed. The resulting thermal images exhibited superior contrast, texture fidelity and structural preservation (Fig.~\ref{fig:case4}\textcolor{blue}{c}).

This case illustrates InfEngine’s capacity for autonomous methodology discovery and adaptive workflow synthesis. Through closed-loop evolution integrating solution generation, automated evaluation and iterative refinement, the system learned to tailor pipelines to both data characteristics and performance objectives. Such capabilities hold transformative potential for multimodal imaging, remote sensing and computational photography, where optimal algorithm selection is inherently instance dependent.

\section{Discussion}\label{sec:discussion}
Infrared radiation computing is gated by the manual effort of generating executable computational workflows. While LLMs have unlocked new forms of automation, existing paradigms fail to provide the reusable, reliable, scientifically grounded and optimized intelligence needed for complex domain-specific computational challenges.

We introduce InfEngine, an intelligent computational engine that fundamentally reimagines this process. By orchestrating a collaborative network of specialized agents for analysis, solving, evaluation and evolution, InfEngine embodies a new paradigm in computational science: the transition from human-driven orchestration of passive tools to collaboration with an autonomous computational partner. Extensive evaluation on InfBench demonstrates that InfEngine significantly outperforms state-of-the-art methods. By generating well-structured, reusable code rather than one-off outputs, InfEngine effectively encapsulates and perpetuates scientific and engineering expertise, turning ad-hoc computational experiments into shareable, extensible research assets.

While InfEngine advances toward a new paradigm for autonomous scientific computing, several limitations highlight clear pathways for future research and development. \emph{First}, the scope of InfTools, while curated for breadth, is necessarily limited compared to the vast, often proprietary toolchains of real-world scientific and engineering practice. A critical next step is to demonstrate how this framework enables domain experts to encapsulate their own specialized tools using the provided interface standard, thereby orchestrating previously isolated capabilities. \emph{Second}, our validation is currently centered on infrared radiation computing. Extending the benchmark and evaluating the core architecture's efficacy in adjacent fields (e.g., computational chemistry, materials informatics) is a vital direction to prove its generalizability. \emph{Finally}, the evolutionary process can be computationally intensive. Future work will explore more efficient search strategies to reduce the optimization cost. Addressing these limitations will transition InfEngine from a powerful prototype into a robust, scalable engine for automated scientific discovery.

Looking forward, the architectural principles of InfEngine point toward an even more transformative future: the evolution of intelligent computational engine into intelligent computational platform. Such a platform would aggregate and orchestrate domain-specific tools, both from curated public repositories and encapsulated private libraries across institutions, through a unified interface standard. Researchers could pose complex problems to this platform, which would then synthesize verified, optimized solver tailored to their specific computational environment and resources. The user would receive and execute this portable code locally, seamlessly bridging the platform's intelligence with private data and hardware. This vision of a federated, tool-agnostic platform transcends individual scientific domains, integrating diverse computational capabilities into a cohesive ecosystem for discovery.

This paradigm shift, from isolated tools to intelligent platforms, has the potential to fundamentally recalibrate the scientific research cycle. By dramatically accelerating the translation of conceptual ideas into executable, validated computational procedures, it augments scientific creativity and productivity. The platform democratizes access to advanced computational methods, liberating researchers from routine implementation burdens and refocusing human intellect on high-level problem formulation, interpretative reasoning, and strategic decision-making---the quintessential elements of breakthrough. We envision a future where such collaborative intelligence becomes a ubiquitous pillar of the scientific method, forming a symbiotic partnership between human intuition and machine execution to accelerate discovery across disciplines.

\section{Method}\label{sec:method}
\subsection{Method Overview}
Our work introduces a new paradigm for autonomous scientific computing in the infrared domain, realized through a multi-agent architecture that translates user problems into verified, optimized and deployable code solutions. As illustrated in Fig.~\ref{fig:overview}, our work is built upon three synergistic and foundational pillars.

\emph{First}, \textbf{InfTools} (Fig.~\ref{fig:overview}\textcolor{blue}{f}) provides the essential computational primitives. This curated suite of 270 domain-specific tools, encapsulated in a standardized format for unified description and invocation (Fig.~\ref{fig:overview}\textcolor{blue}{g}), constitutes the executable knowledge base that grounds InfEngine in the infrared radiation computing domain. \emph{Second}, \textbf{InfBench} establishes the rigorous evaluation ground. Comprising 200 well-defined tasks across assistant and optimization types, this benchmark supplies the diverse and challenging scenarios necessary to validate the autonomous capabilities of InfEngine, ensuring its robustness and generality. \emph{Third}, the core innovation is \textbf{InfEngine} itself (Fig.~\ref{fig:overview}\textcolor{blue}{e}), the intelligent computational engine proposed in this work. Users interact with InfEngine as a computational partner, posing complex problems in natural language. The system orchestrates a collaborative network of specialized agents to analyze, plan, solve, evaluate, and iteratively optimize the problem by dynamically integrating tools from both private tools and public libraries (ref. Fig.~\ref{fig:overview}\textcolor{blue}{h}). Critically, its final output is not a transient answer but a production-ready, reusable solver code delivered to the user. This addresses a key limitation of the tool-calling paradigm, which typically yields only specific, non-reusable outputs. Our approach provides a generalizable and executable solution procedure that encapsulates the entire solved workflow.

In summary, the proposed InfEngine provides a unified solution to the diverse computational challenges across the infrared radiation computing pipeline, as depicted in Fig.~\ref{fig:overview}\textcolor{blue}{b}. It automates the process from problem description to the generation of validated and optimized code, thereby powering the wide range of applications in domains shown in Fig.~\ref{fig:overview}\textcolor{blue}{a}.

\subsection{InfTools}
The creation of a comprehensive, machine-executable knowledge base is foundational to InfEngine’s code generation paradigm. We present \textbf{InfTools}, a systematically curated collection of \textbf{270} interoperable tools for infrared radiation computing. To our knowledge, this represents the first large-scale assembly of computational resources specifically designed for LLM-driven automation in this domain. While broader collections exist for other sciences (e.g., SciToolAgent's~\cite{ding2025scitoolagent} ~500 tools for biology and chemistry, ChemCrow~\cite{ChemCrow} and ChatMOF~\cite{ChatMOF} operate with 18 and 4 tools respectively for chemical problems), InfTools is distinguished by its exclusive domain focus and its design for seamless integration into generated code, contrasting with frameworks built for rigid tool invocation.

\bmhead{Tool Collection and Curation}Adhering to principles of reproducibility and open science, we sourced tools exclusively from public repositories (primarily GitHub), using targeted queries (“radiative transfer”, “infrared spectrum analysis”, etc.). Candidate codebases underwent stringent filtering to exclude projects with incomplete implementations, missing trained models, or obsolete dependencies, ensuring a robust foundation.

\bmhead{Standardized Tool Encapsulation}Each tool was encapsulated using a rigorous, consistent schema capturing: \textbf{name}, \textbf{description}, \textbf{input/output specification} (formatted as ``type - explanation''), \textbf{usage examples}, \textbf{dependencies}, \textbf{source link}, \textbf{build command} (a shell script to reconstruct the runnable package from source), and \textbf{user metadata} (limitations, related papers). Crucially, our schema supports rich output types, including strings, NumPy arrays, and class instances, which overcomes a common restriction in tool-calling frameworks that only permit hashable types.

Encapsulation employed a tiered strategy based on complexity. For simple scripts, an LLM-assisted extraction process efficiently isolated core logic. For complex, multi-file projects, manual refactoring was necessary to produce clean, modular Python interfaces. This process transformed disparate codebases into a unified, callable library.

\bmhead{Validation and Quality Assurance}Each encapsulated tool was validated against original functionality through a battery of test cases with known expected outputs. Discrepancies triggered iterative debugging until computational fidelity was guaranteed. This labor-intensive process, requiring approximately \textbf{five person-months} of expert effort, underscores the significant investment needed to create production-ready, trustworthy resources for AI-driven science. The resulting InfTools suite provides a reliable, scalable substrate that enables InfEngine to reason over and compose complex infrared radiation computing workflows.

\subsection{InfBench}
To rigorously evaluate InfEngine’s capabilities, we constructed \textbf{InfBench}, a domain-specific benchmark with a key innovation in its dual problem taxonomy. This taxonomy differentiates between two core task types central to our study: \textit{assistant-type} tasks, which require correct, one-pass solutions, and \textit{optimization-type} tasks, which demand iterative refinement to meet user-specified objectives. In total, InfBench comprises 200 distinct test tasks, split into 165 assistant-type tasks and 35 optimization-type tasks. This tailored design directly addresses a critical gap in existing benchmark evaluations such as GAIA~\cite{GAIA}, which overwhelmingly prioritize the assessment of assistant-style capabilities while largely neglecting the optimization competencies essential for advanced scientific computing scenarios.

Each task in InfBench is a structured quadruplet: (\textbf{i}) a natural language question describing the computational problem; (\textbf{ii}) a \texttt{solve(tools, **kwargs)} function that provides a reference solver, where \texttt{tools} is a dictionary of instantiated tool objects accessible during execution, and \texttt{kwargs} are keyword arguments representing the problem's default input parameters; (\textbf{iii}) a candidate tool list enumerating all tools from the InfTools repository that are potentially applicable, which serves as the basis for evaluating an agent's tool selection accuracy; and (\textbf{iv}) a standardized \texttt{evaluate(result, **kwargs)} function for solver assessment. For assistant-type tasks, \texttt{evaluate} returns a binary score (1 for success, 0 for failure), typically implemented via assertions. For optimization-type tasks, it returns a normalized quality score (higher values indicate better solutions), scaled to concentrate near 1 for consistent cross-task comparison. The \texttt{evaluate} function may itself call upon tools from InfTools, enabling complex, domain-aware validation.

InfBench was constructed entirely through manual curation by domain experts to ensure the high standards of quality, correctness, and relevance. For each task, experts performed the following steps: \emph{First}, they formulated a natural language question that captures a realistic computational challenge related to infrared radiation computing. \emph{Second}, they manually implemented the reference \texttt{solve} function, ensuring it correctly utilizes the appropriate tools from InfTools to produce a valid solution. \emph{Third}, they curated the candidate tool list, carefully selecting tools that are both relevant and plausible for solving the given problem. \emph{Fourth}, they authored the \texttt{evaluate} function, embedding precise success criteria for assistant-type tasks or designing robust scoring rubrics for optimization-type tasks. This rigorous, hands-on process guaranteed that every benchmark task is well-specified, executable and grounded in authentic domain expertise.

Constructing InfBench required a substantial investment of expert effort, with approximately \textbf{1.5 hours} devoted to the manual creation and validation of each task. This deliberate approach prioritizes benchmark reliability and fidelity over scale, resulting in a high-quality evaluation framework. The resulting benchmark, with its explicit tool lists and standardized dual-mode evaluation, enables precise measurement of an AI system's ability to select appropriate tools, generate functionally correct code, and optimize solutions against quantifiable goals, providing the comprehensive framework needed to validate scientific computing agents.

\subsection{InfEngine}
The InfEngine materializes the two core capabilities outlined in the introduction: \textbf{self‑verification} and \textbf{self‑optimization}. As shown in Fig.~\ref{fig:overview}\textcolor{blue}{e}, it is orchestrated by a multi‑agent architecture in which specialized agents collaborate to transform a user’s natural‑language problem into verified and optimized code. The workflow begins with a \textbf{Problem Analyzing Agent} that classifies the query and retrieves relevant tools. For an assistant‑type task, a \textbf{Problem Solving Agent} and an \textbf{Evaluator Generation Agent} work in tandem to achieve self‑verification, producing executable code that is jointly debugged against a context‑aware evaluation script. For an optimization problem, the verified solution is passed to a \textbf{Code Evolution Agent} to perform self‑optimization, iteratively refining the code against the auto‑generated evaluator. The final output is a standalone, packaged solution with all dependencies.

\subsubsection{Problem Analyzing Agent}
The Problem Analyzing Agent initiates the workflow by interpreting the natural‑language problem and structuring it into a well‑defined specification. Its core task is to formalize an ambiguous user query into a clear specification comprising three key components: the formal task description $T$, the optimization goal $G$, and a reference answer $A_{\text{ref}}$. This separation is crucial: $T$ defines what the solver must do, $G$ provides the metric for how well it should perform, and $A_{\text{ref}}$ serves as a verification standard. Crucially, isolating $A_{\text{ref}}$ prevents potential cheating by ensuring it is used only for final validation and not prematurely revealed during the solution process.

Formally, the agent maps a natural‑language query $q$ to a structured specification tuple $(T, G, A_{\text{ref}})$ and a set of relevant tools $\mathcal{T}_{\text{solve}}$:
\[
\mathcal{A}_{\text{ana}}: q \mapsto \big( (T, G, A_{\text{ref}}), \; \mathcal{T}_{\text{solve}} \big).
\]

The agent first classifies $q$ into a task type $\tau \in \{\text{assist}, \text{opt}\}$ using an LLM with a classification prompt $p_{\text{cls}}$:
\[
\tau = \operatorname{LLM}(q; p_{\text{cls}}).
\]

For assistant‑type tasks ($\tau = \text{assist}$), the specification is defined directly from the original query:
\[
T = q, \quad G = \varnothing, \quad A_{\text{ref}} = \varnothing.
\]

For optimization‑type tasks ($\tau = \text{opt}$), the agent parses $q$ into five structured elements using a formalization prompt $p_{\text{fm}}$:
\[
\{I_{\text{in}}, I_{\text{out}}, I_{\text{inst}}, g_{\text{raw}}, a_{\text{gt}}\} = \operatorname{LLM}(q; p_{\text{fm}}),
\]
where $I_{\text{in}}$ denotes the input description, $I_{\text{out}}$ the output format, $I_{\text{inst}}$ the step‑by‑step instructions, $g_{\text{raw}}$ the explicit optimization goal, and $a_{\text{gt}}$ the ground‑truth answer. These elements are then programmatically synthesized into the final three‑component specification:
\[
T = \text{``Inputs: }I_{\text{in}}\text{ Output Format: }I_{\text{out}}\text{ Instructions: }I_{\text{inst}}\text{''}, \quad
G = g_{\text{raw}}, \quad
A_{\text{ref}} = a_{\text{gt}}.
\]
This decompose‑then‑recompose approach proves more reliable than direct task generation, as it ensures each component is explicitly defined.

Concurrently, the agent retrieves a relevant set of candidate tools from the InfTools repository using a RAG strategy. The query $q$ is encoded into a dense vector $\mathbf{e}_q \in \mathbb{R}^{768}$ using the \textit{all‑mpnet‑base‑v2} sentence‑transformer model:
\[
\mathbf{e}_q = \operatorname{Encoder}(q).
\]
For each tool description $d_j$ in the database with pre‑computed embedding $\mathbf{d}_j$, the similarity score is computed as the cosine similarity:
\[
s_j = \cos(\mathbf{e}_q, \mathbf{d}_j).
\]
The top‑$k$ tools with the highest scores are retrieved to form the candidate set:
\[
\mathcal{T}_{\text{solve}} = \underset{j \in [1, M]}{\operatorname{top-k}} \; s_j,
\]
where $M$ is the total number of  tools.

\subsubsection{Problem Solving Agent}
The Problem Solving Agent translates the formalized task description $T$ into executable code through an iterative process of planning, generation, execution and validation. Formally, the agent maps the task and retrieved tools to an executable code:
\[
\mathcal{A}_{\text{sol}}: (T, \mathcal{T}_{\text{solve}}) \mapsto c_{\text{code}}.
\]

The process begins with a planning phase, where the agent decomposes $T$ into a structured plan $\pi$ using a planning prompt $p_{\text{plan}}$:
\[
\pi = \operatorname{LLM}(T, \mathcal{T}_{\text{solve}}; p_{\text{plan}}),
\]
where the plan $\pi$ is a JSON structure specifying sub‑tasks, required tools, Python packages, and a rationale for the decomposition.

Using a predefined code template $\tau_{\text{tmpl}}$, the agent then generates the initial solution code by instantiating the plan within the template with a generation prompt $p_{\text{gen}}$:
\[
c_0 = \operatorname{LLM}(T, \mathcal{T}_{\text{solve}}, \pi, \tau_{\text{tmpl}}; p_{\text{gen}}).
\]

The generated code is executed in a strictly controlled Python environment. Let the execution function be $\operatorname{Exec}(\cdot)$. The result and any error are obtained as:
\[
r_i, \; \varepsilon_i = \operatorname{Exec}(c_i).
\]
If an error occurs ($\varepsilon_i \neq \varnothing$), a debugging step is invoked to produce a corrected version using a debugging prompt $p_{\text{dbg}}$:
\[
c_{i+1} = \operatorname{LLM}(T, \mathcal{T}_{\text{solve}}, c_i, \varepsilon_i; p_{\text{dbg}}).
\]
This execution‑debugging loop repeats until either (1) the code runs successfully ($\varepsilon_i = \varnothing$), or (2) a maximum debugging iteration limit is reached. If the debugging limit is reached without producing error‑free code, the entire process reverts to the planning phase to generate a fundamentally new solution strategy.

Upon successful execution, the result $r_{\text{res}}$ is validated by an evaluator $E$ generated by the Evaluator Generation Agent. The validation yields a score $s$ that quantifies correctness or quality:
\[
s = E(r_{\text{res}}, A_{\text{ref}}).
\]
For assistant‑type tasks, the score is binary, $s \in \{0,1\}$; for optimization-type tasks, it is a normalized continuous value, $s \in [0,1]$. If the above validation process meets failure, the entire cycle is re‑initiated unless a maximum iteration limit is reached.

\subsubsection{Evaluator Generation Agent}
The Evaluator Generation Agent produces a robust evaluation function that enables self‑verification. Formally, it constructs an evaluator $E$ from the problem specification $(T, G, A_\text{ref})$, the available evaluation-related tools $\mathcal{T}_\text{eval}$, the candidate solution $c_{\text{code}}$, and its execution result $r_{\text{res}}$:
\[
\mathcal{A}_{\text{eval}}: (T, G, A_{\text{ref}}, \mathcal{T}_\text{eval}, c_{\text{code}}, r_{\text{res}}) \mapsto E.
\]

For optimization-type tasks, the agent first employs a planner to decompose the validation objective into manageable sub‑tasks. Using a planning prompt $p_{\text{plan\_eval}}$, it produces a structured validation plan $\pi_{\text{eval}}$:
\[
\pi_{\text{eval}} = \operatorname{LLM}(T, G, A_{\text{ref}}, \mathcal{T}_\text{eval}, c_{\text{code}}; p_{\text{plan\_eval}}).
\]
This plan is then instantiated into executable evaluator code with a generation prompt $p_{\text{gen\_eval}}$:
\[
E_{\text{init}} = \operatorname{LLM}(T, G, A_{\text{ref}}, \mathcal{T}_\text{solve}, \mathcal{T}_\text{eval}, c_{\text{code}}, \pi_{\text{eval}}; p_{\text{gen\_eval}}).
\]

For assistant-type tasks, whose verification logic is typically simpler, the planning phase is omitted, and the evaluator is generated directly from the output format using a dedicated prompt $p_{\text{gen\_eval\_assist}}$:
\[
E_{\text{init}} = \operatorname{LLM}(T, \mathcal{T}_\text{solve}, \mathcal{T}_\text{eval}, c_{\text{code}}; p_{\text{gen\_eval\_assist}}).
\]

A critical innovation is the joint solver‑evaluator debugging mechanism. If the execution of $E_{\text{init}}$ raises an error $\varepsilon_E$ while evaluating the current solver $c_\text{code}$, a dedicated referee module performs a dual diagnostic to isolate the root cause. The referee determines whether the error originates from the solver’s implementation, the evaluator’s rule specification, or an ill‑defined interaction between them. Guided by a carefully designed prompt $p_{\text{ref}}$, the referee produces corrected versions of both the solution code and the evaluator:
\[
(c_{\text{code}}', E') = \operatorname{LLM}(T, c_{\text{code}}, r_{\text{res}}, \mathcal{T}_\text{solve}, \mathcal{T}_\text{eval}, E_{\text{init}}, \varepsilon_E; p_{\text{ref}}).
\]
Throughout this correction, integrity constraints are applied to prevent spurious co-optimization, enforcing that modifications preserve the solver's generalizability and the evaluator's fidelity to the original problem constraints. The process iterates until the evaluator executes successfully or maximum iteration limit is reached. This iterative debugging process helps to improve the solver's functional correctness and scientific plausibility.

\subsubsection{Code Evolution Agent}
The Code Evolution Agent introduces the self-optimization capability by autonomously refining solution codes through an evolutionary process. Formally, the agent evolves an initial code $c$ over $N$ generations, using the auto-generated evaluator $E$ as a fitness function, to produce an optimized solution:
\[
\mathcal{A}_{\text{evol}}: (c, E, N) \mapsto c^{*}.
\]

The fitness of a code individual $c$ is defined as the normalized score returned by the evaluator on its execution result:
\[
f(c) = E(\operatorname{Exec}(c), A_\text{ref}).
\]

At each generation $t$, the agent maintains a population $P_t$ whose size may vary during the evolutionary cycle. Initially, the population size is $1$. Let $c^{(t)}_r$ denote the individual with rank $r$ in generation $t$, where individuals are sorted by fitness value. Rank-based selection probabilities are computed to favor higher-ranked individuals while maintaining diversity:
\[
P(c^{(t)}_r) = \frac{1/(r + |P_t|)}{\sum_{i=1}^{|P_t|} 1/(i + |P_t|)},
\]
where $r \in \{1, \dots, |P_t|\}$ denotes the rank of the individual (with rank 1 for the highest fitness).

Selected parents undergo variation through a suite of genetically inspired operators. Each operator works by prompting the LLM to perform strategic reasoning that produces both an algorithmic description and its implementation in a single step. The LLM generates a blueprint, a high-level description of the algorithm's strategy, and instantiates this blueprint into executable code, guided by the system's structural template.

\textbf{Crossover operators} (E1 and E2) synthesize novel algorithms by recombining multiple parent solutions. Let $\{c_{\text{parents}}\} \subset P_t$ denote the set of parent code individuals selected from the current population. Each prompt instructs the LLM to first articulate the core algorithmic idea (or, for E2, to identify a common backbone idea), and then to implement it. The output is a unified response containing both the strategic description and the runnable code:
\[
(\text{description}, c_{\text{child}}) = \operatorname{LLM}(T, \mathcal{T}_\text{solve}, \{c_{\text{parents}}\}, \tau_{\text{tmpl}}; p_{\text{cross}}).
\]

\textbf{Mutation operators} (M1, M2, M3) generate variants from a single parent, each targeting a different level of modification. Let $c_{\text{parent}} \in P_t$ denote the parent code individual selected from the current population for mutation. The prompts similarly require the LLM to first describe the modified strategy (e.g., architectural change, tool reconfiguration, or parameter tuning) and then produce the corresponding code. This single-step generation guarantees that the strategic blueprint is faithfully translated into a syntactically correct solution:
\[
(\text{description}, c_{\text{mut}}) = \operatorname{LLM}(T, \mathcal{T}_\text{solve}, c_{\text{parent}}, \tau_{\text{tmpl}}; p_{\text{mut}}).
\]

All operators thus operate at the strategic level, using the algorithmic description as a blueprint for immediate code generation. The output code is structured according to the predefined template $\tau_{\text{tmpl}}$, ensuring interface compatibility.

After evaluating the fitness of all offspring, the population is managed to preserve quality and diversity. Invalid individuals with undefined fitness are removed. Duplicate solutions with identical fitness values are deduplicated. Finally, elitist selection retains at most $m$ individuals with the highest fitness to form the next generation, ensuring the population size does not exceed the predefined limit.

This cycle of selection, variation, evaluation and population management repeats for $N$ generations. The algorithm converges by returning the highest-fitness individual from the final population:
\[
c^{*} = \underset{c \in P_N}{\arg\max} \; f(c).
\]

Through this evolutionary process, the system autonomously discovers and optimizes computational solutions against a machine-synthesized objective, closing the loop from informal problem description to high-performance code artifact.

\subsubsection{Project Construction}
Upon completing the core multi-agent workflow, InfEngine can further package the validated solution into a directly deployable and reusable software module. The process begins with a dependency analysis of the final solution code to identify all required tools from the InfTools repository. The system then copies these tools along with their associated resource files into a structured project directory. Next, it aggregates dependency specifications from all involved tools to generate comprehensive environment configuration files, ensuring the solution can be reliably reproduced across different computing environments. The validated solve function is then packaged as an installable Python module. Through automated code restructuring, appropriate metadata, API documentation and entry points are added.

Through this project construction process, InfEngine yields not just executable code but a reusable artifact: each packaged \texttt{solve} function constitutes a new, stand-alone composite tool. This artifact embodies a dual pathway to reusability. \emph{First}, from the user's perspective, the delivered solver is not merely a terminal answer but a modifiable and integrable software component. Users can directly deploy it, inspect its logic, adapt it to new inputs, or embed it within their own projects. \emph{Second}, from the system's perspective, these synthesized solvers can themselves be formalized and added back to the computational tool repository. This creates a virtuous cycle: InfEngine not only consumes a foundational set of tools but also produces higher-level, task-specific tools, progressively enriching its own capabilities and reducing the complexity of solving future, related problems. This self-enhancing mechanism transforms the paradigm from a static toolbox to a dynamic, learning ecosystem that continuously expands the frontier of automated computational science.

\section{Acknowledgements}
This work was supported by the Strategic Priority Research Program of Chinese Academy of Sciences (Grant No. XDA0480200) and the National Natural Science Foundations of China (Grant No.62306310).

\section{Competing Interests}
The authors declare no competing interests.

\section{Data Availability}
The infrared radiation computing benchmark (InfBench) dataset supporting the findings of this study is openly available in the repository: \url{https://github.com/kding1225/infengine}.

\section{Code Availability}
The InfEngine framework and its associated toolset are publicly available as open-source software in the repository at \url{https://github.com/kding1225/infengine}.


\bibliography{sn-bibliography}


\begin{thebibliography}{32}
\ifx \bisbn   \undefined \def \bisbn  #1{ISBN #1}\fi
\ifx \binits  \undefined \def \binits#1{#1}\fi
\ifx \bauthor  \undefined \def \bauthor#1{#1}\fi
\ifx \batitle  \undefined \def \batitle#1{#1}\fi
\ifx \bjtitle  \undefined \def \bjtitle#1{#1}\fi
\ifx \bvolume  \undefined \def \bvolume#1{\textbf{#1}}\fi
\ifx \byear  \undefined \def \byear#1{#1}\fi
\ifx \bissue  \undefined \def \bissue#1{#1}\fi
\ifx \bfpage  \undefined \def \bfpage#1{#1}\fi
\ifx \blpage  \undefined \def \blpage #1{#1}\fi
\ifx \burl  \undefined \def \burl#1{\textsf{#1}}\fi
\ifx \doiurl  \undefined \def \doiurl#1{\url{https://doi.org/#1}}\fi
\ifx \betal  \undefined \def \betal{\textit{et al.}}\fi
\ifx \binstitute  \undefined \def \binstitute#1{#1}\fi
\ifx \binstitutionaled  \undefined \def \binstitutionaled#1{#1}\fi
\ifx \bctitle  \undefined \def \bctitle#1{#1}\fi
\ifx \beditor  \undefined \def \beditor#1{#1}\fi
\ifx \bpublisher  \undefined \def \bpublisher#1{#1}\fi
\ifx \bbtitle  \undefined \def \bbtitle#1{#1}\fi
\ifx \bedition  \undefined \def \bedition#1{#1}\fi
\ifx \bseriesno  \undefined \def \bseriesno#1{#1}\fi
\ifx \blocation  \undefined \def \blocation#1{#1}\fi
\ifx \bsertitle  \undefined \def \bsertitle#1{#1}\fi
\ifx \bsnm \undefined \def \bsnm#1{#1}\fi
\ifx \bsuffix \undefined \def \bsuffix#1{#1}\fi
\ifx \bparticle \undefined \def \bparticle#1{#1}\fi
\ifx \barticle \undefined \def \barticle#1{#1}\fi
\bibcommenthead
\ifx \bconfdate \undefined \def \bconfdate #1{#1}\fi
\ifx \botherref \undefined \def \botherref #1{#1}\fi
\ifx \url \undefined \def \url#1{\textsf{#1}}\fi
\ifx \bchapter \undefined \def \bchapter#1{#1}\fi
\ifx \bbook \undefined \def \bbook#1{#1}\fi
\ifx \bcomment \undefined \def \bcomment#1{#1}\fi
\ifx \oauthor \undefined \def \oauthor#1{#1}\fi
\ifx \citeauthoryear \undefined \def \citeauthoryear#1{#1}\fi
\ifx \endbibitem  \undefined \def \endbibitem {}\fi
\ifx \bconflocation  \undefined \def \bconflocation#1{#1}\fi
\ifx \arxivurl  \undefined \def \arxivurl#1{\textsf{#1}}\fi
\csname PreBibitemsHook\endcsname

\bibitem[\protect\citeauthoryear{Yin et~al.}{2018}]{yin2018ultrafast}
\begin{barticle}
\bauthor{\bsnm{Yin}, \binits{J.}},
\bauthor{\bsnm{Tan}, \binits{Z.}},
\bauthor{\bsnm{Hong}, \binits{H.}},
\bauthor{\bsnm{Wu}, \binits{J.}},
\bauthor{\bsnm{Yuan}, \binits{H.}},
\bauthor{\bsnm{Liu}, \binits{Y.}},
\bauthor{\bsnm{Chen}, \binits{C.}},
\bauthor{\bsnm{Tan}, \binits{C.}},
\bauthor{\bsnm{Yao}, \binits{F.}},
\bauthor{\bsnm{Li}, \binits{T.}}, \betal:
\batitle{Ultrafast and highly sensitive infrared photodetectors based on
  two-dimensional oxyselenide crystals}.
\bjtitle{Nature communications}
\bvolume{9}(\bissue{1}),
\bfpage{3311}
(\byear{2018})
\end{barticle}
\endbibitem

\bibitem[\protect\citeauthoryear{Cheng et~al.}{2022}]{cheng20222d}
\begin{barticle}
\bauthor{\bsnm{Cheng}, \binits{Z.}},
\bauthor{\bsnm{Zhao}, \binits{T.}},
\bauthor{\bsnm{Zeng}, \binits{H.}}:
\batitle{2d material-based photodetectors for infrared imaging}.
\bjtitle{Small Science}
\bvolume{2}(\bissue{1}),
\bfpage{2100051}
(\byear{2022})
\end{barticle}
\endbibitem

\bibitem[\protect\citeauthoryear{Alberts et~al.}{2024}]{alberts2024leveraging}
\begin{barticle}
\bauthor{\bsnm{Alberts}, \binits{M.}},
\bauthor{\bsnm{Laino}, \binits{T.}},
\bauthor{\bsnm{Vaucher}, \binits{A.C.}}:
\batitle{Leveraging infrared spectroscopy for automated structure elucidation}.
\bjtitle{Commun. Chem.}
\bvolume{7},
\bfpage{268}
(\byear{2024})
\end{barticle}
\endbibitem

\bibitem[\protect\citeauthoryear{Ottomano et~al.}{2025}]{NMIRacle}
\begin{botherref}
\oauthor{\bsnm{Ottomano}, \binits{F.}},
\oauthor{\bsnm{Li}, \binits{Y.}},
\oauthor{\bsnm{Ganose}, \binits{A.M.}}:
NMIRacle: Multi-modal Generative Molecular Elucidation from IR and NMR Spectra.
Preprint at \url{https://arxiv.org/abs/2512.19733}
(2025)
\end{botherref}
\endbibitem

\bibitem[\protect\citeauthoryear{Clough et~al.}{2005}]{CLOUGH2005233}
\begin{barticle}
\bauthor{\bsnm{Clough}, \binits{S.A.}},
\bauthor{\bsnm{Shephard}, \binits{M.W.}},
\bauthor{\bsnm{Mlawer}, \binits{E.J.}},
\bauthor{\bsnm{Delamere}, \binits{J.S.}},
\bauthor{\bsnm{Iacono}, \binits{M.J.}},
\bauthor{\bsnm{Cady-Pereira}, \binits{K.}},
\bauthor{\bsnm{Boukabara}, \binits{S.}},
\bauthor{\bsnm{Brown}, \binits{P.D.}}:
\batitle{Atmospheric radiative transfer modeling: a summary of the aer codes}.
\bjtitle{Journal of Quantitative Spectroscopy and Radiative Transfer}
\bvolume{91}(\bissue{2}),
\bfpage{233}--\blpage{244}
(\byear{2005})
\end{barticle}
\endbibitem

\bibitem[\protect\citeauthoryear{Wilson et~al.}{2023}]{10014654}
\begin{barticle}
\bauthor{\bsnm{Wilson}, \binits{A.N.}},
\bauthor{\bsnm{Gupta}, \binits{K.A.}},
\bauthor{\bsnm{Koduru}, \binits{B.H.}},
\bauthor{\bsnm{Kumar}, \binits{A.}},
\bauthor{\bsnm{Jha}, \binits{A.}},
\bauthor{\bsnm{Cenkeramaddi}, \binits{L.R.}}:
\batitle{Recent advances in thermal imaging and its applications using machine
  learning: A review}.
\bjtitle{IEEE Sensors Journal}
\bvolume{23}(\bissue{4}),
\bfpage{3395}--\blpage{3407}
(\byear{2023})
\end{barticle}
\endbibitem

\bibitem[\protect\citeauthoryear{Leroy
  et~al.}{2025}]{Leroy_Eradiate_radiative_transfer_2025}
\begin{botherref}
\oauthor{\bsnm{Leroy}, \binits{V.}},
\oauthor{\bsnm{Nollet}, \binits{Y.}},
\oauthor{\bsnm{Schunke}, \binits{S.}},
\oauthor{\bsnm{Misk}, \binits{N.}},
\oauthor{\bsnm{Marton}, \binits{N.}},
\oauthor{\bsnm{Emde}, \binits{C.}},
\oauthor{\bsnm{Govaerts}, \binits{Y.}}:
Eradiate Radiative Transfer Model.
\url{https://github.com/eradiate/eradiate}.
Software available at \url{https://github.com/eradiate/eradiate}
(2025)
\end{botherref}
\endbibitem

\bibitem[\protect\citeauthoryear{Emde et~al.}{2016}]{emde2016libradtran}
\begin{barticle}
\bauthor{\bsnm{Emde}, \binits{C.}},
\bauthor{\bsnm{Buras-Schnell}, \binits{R.}},
\bauthor{\bsnm{Kylling}, \binits{A.}},
\bauthor{\bsnm{Mayer}, \binits{B.}},
\bauthor{\bsnm{Gasteiger}, \binits{J.}},
\bauthor{\bsnm{Hamann}, \binits{U.}},
\bauthor{\bsnm{Kylling}, \binits{J.}},
\bauthor{\bsnm{Richter}, \binits{B.}},
\bauthor{\bsnm{Pause}, \binits{C.}},
\bauthor{\bsnm{Dowling}, \binits{T.}}, \betal:
\batitle{The libradtran software package for radiative transfer calculations
  (version 2.0. 1)}.
\bjtitle{Geoscientific Model Development}
\bvolume{9}(\bissue{5}),
\bfpage{1647}--\blpage{1672}
(\byear{2016})
\end{barticle}
\endbibitem

\bibitem[\protect\citeauthoryear{Du and Stechmann}{2023}]{du2023fast}
\begin{barticle}
\bauthor{\bsnm{Du}, \binits{S.}},
\bauthor{\bsnm{Stechmann}, \binits{S.N.}}:
\batitle{Fast, low-memory numerical methods for radiative transfer via
  hp-adaptive mesh refinement}.
\bjtitle{Journal of Computational Physics}
\bvolume{480},
\bfpage{112021}
(\byear{2023})
\end{barticle}
\endbibitem

\bibitem[\protect\citeauthoryear{Marzano}{2014}]{Marzano2014}
\begin{bbook}
\bauthor{\bsnm{Marzano}, \binits{F.S.}}:
In: \beditor{\bsnm{Njoku}, \binits{E.G.}} (ed.)
\bbtitle{Radiative Transfer, Theory},
pp. \bfpage{624}--\blpage{634}.
\bpublisher{Springer},
\blocation{New York, NY}
(\byear{2014})
\end{bbook}
\endbibitem

\bibitem[\protect\citeauthoryear{Brown et~al.}{2020}]{NEURIPS2020_1457c0d6}
\begin{bchapter}
\bauthor{\bsnm{Brown}, \binits{T.}},
\bauthor{\bsnm{Mann}, \binits{B.}},
\bauthor{\bsnm{Ryder}, \binits{N.}},
\bauthor{\bsnm{Subbiah}, \binits{M.}},
\bauthor{\bsnm{Kaplan}, \binits{J.D.}},
\bauthor{\bsnm{Dhariwal}, \binits{P.}},
\bauthor{\bsnm{Neelakantan}, \binits{A.}},
\bauthor{\bsnm{Shyam}, \binits{P.}},
\bauthor{\bsnm{Sastry}, \binits{G.}},
\bauthor{\bsnm{Askell}, \binits{A.}},
\bauthor{\bsnm{Agarwal}, \binits{S.}},
\bauthor{\bsnm{Herbert-Voss}, \binits{A.}},
\bauthor{\bsnm{Krueger}, \binits{G.}},
\bauthor{\bsnm{Henighan}, \binits{T.}},
\bauthor{\bsnm{Child}, \binits{R.}},
\bauthor{\bsnm{Ramesh}, \binits{A.}},
\bauthor{\bsnm{Ziegler}, \binits{D.}},
\bauthor{\bsnm{Wu}, \binits{J.}},
\bauthor{\bsnm{Winter}, \binits{C.}},
\bauthor{\bsnm{Hesse}, \binits{C.}},
\bauthor{\bsnm{Chen}, \binits{M.}},
\bauthor{\bsnm{Sigler}, \binits{E.}},
\bauthor{\bsnm{Litwin}, \binits{M.}},
\bauthor{\bsnm{Gray}, \binits{S.}},
\bauthor{\bsnm{Chess}, \binits{B.}},
\bauthor{\bsnm{Clark}, \binits{J.}},
\bauthor{\bsnm{Berner}, \binits{C.}},
\bauthor{\bsnm{McCandlish}, \binits{S.}},
\bauthor{\bsnm{Radford}, \binits{A.}},
\bauthor{\bsnm{Sutskever}, \binits{I.}},
\bauthor{\bsnm{Amodei}, \binits{D.}}:
\bctitle{Language models are few-shot learners}.
In: \bbtitle{Advances in Neural Information Processing Systems},
vol. \bseriesno{33},
pp. \bfpage{1877}--\blpage{1901}
(\byear{2020})
\end{bchapter}
\endbibitem

\bibitem[\protect\citeauthoryear{Qu et~al.}{2025}]{QuDWCWYXW25}
\begin{barticle}
\bauthor{\bsnm{Qu}, \binits{C.}},
\bauthor{\bsnm{Dai}, \binits{S.}},
\bauthor{\bsnm{Wei}, \binits{X.}},
\bauthor{\bsnm{Cai}, \binits{H.}},
\bauthor{\bsnm{Wang}, \binits{S.}},
\bauthor{\bsnm{Yin}, \binits{D.}},
\bauthor{\bsnm{Xu}, \binits{J.}},
\bauthor{\bsnm{Wen}, \binits{J.}}:
\batitle{Tool learning with large language models: a survey}.
\bjtitle{Frontiers Comput. Sci.}
\bvolume{19}(\bissue{8}),
\bfpage{198343}
(\byear{2025})
\end{barticle}
\endbibitem

\bibitem[\protect\citeauthoryear{Lu et~al.}{2025}]{lu2025octotools}
\begin{bchapter}
\bauthor{\bsnm{Lu}, \binits{P.}},
\bauthor{\bsnm{Chen}, \binits{B.}},
\bauthor{\bsnm{Liu}, \binits{S.}},
\bauthor{\bsnm{Thapa}, \binits{R.}},
\bauthor{\bsnm{Boen}, \binits{J.}},
\bauthor{\bsnm{Zou}, \binits{J.}}:
\bctitle{Octotools: An agentic framework with extensible tools for complex
  reasoning}.
In: \bbtitle{ICLR 2025 Workshop on Foundation Models in the Wild}
(\byear{2025})
\end{bchapter}
\endbibitem

\bibitem[\protect\citeauthoryear{Ma et~al.}{2024}]{SciAgent}
\begin{bchapter}
\bauthor{\bsnm{Ma}, \binits{Y.}},
\bauthor{\bsnm{Gou}, \binits{Z.}},
\bauthor{\bsnm{Hao}, \binits{J.}},
\bauthor{\bsnm{Xu}, \binits{R.}},
\bauthor{\bsnm{Wang}, \binits{S.}},
\bauthor{\bsnm{Pan}, \binits{L.}},
\bauthor{\bsnm{Yang}, \binits{Y.}},
\bauthor{\bsnm{Cao}, \binits{Y.}},
\bauthor{\bsnm{Sun}, \binits{A.}}:
\bctitle{{S}ci{A}gent: Tool-augmented language models for scientific
  reasoning}.
In: \bbtitle{Proceedings of the 2024 Conference on Empirical Methods in Natural
  Language Processing},
\bconflocation{Miami, Florida, USA},
pp. \bfpage{15701}--\blpage{15736}
(\byear{2024})
\end{bchapter}
\endbibitem

\bibitem[\protect\citeauthoryear{Lyu et~al.}{2025}]{lyu2025adapting}
\begin{bchapter}
\bauthor{\bsnm{Lyu}, \binits{B.}},
\bauthor{\bsnm{Cao}, \binits{Y.}},
\bauthor{\bsnm{Watson-Parris}, \binits{D.}},
\bauthor{\bsnm{Bergen}, \binits{L.}},
\bauthor{\bsnm{Berg-Kirkpatrick}, \binits{T.}},
\bauthor{\bsnm{Yu}, \binits{R.}}:
\bctitle{Adapting while learning: Grounding {LLM}s for scientific problems with
  tool usage adaptation}.
In: \bbtitle{Forty-second International Conference on Machine Learning}
(\byear{2025})
\end{bchapter}
\endbibitem

\bibitem[\protect\citeauthoryear{Bran et~al.}{2024}]{ChemCrow}
\begin{barticle}
\bauthor{\bsnm{Bran}, \binits{A.M.}},
\bauthor{\bsnm{Cox}, \binits{S.}},
\bauthor{\bsnm{Schilter}, \binits{O.}},
\bauthor{\bsnm{Baldassari}, \binits{C.}},
\bauthor{\bsnm{White}, \binits{A.D.}},
\bauthor{\bsnm{Schwaller}, \binits{P.}}:
\batitle{Augmenting large language models with chemistry tools}.
\bjtitle{Nat. Mac. Intell.}
\bvolume{6}(\bissue{5}),
\bfpage{525}--\blpage{535}
(\byear{2024})
\end{barticle}
\endbibitem

\bibitem[\protect\citeauthoryear{Kang and Kim}{2024}]{ChatMOF}
\begin{barticle}
\bauthor{\bsnm{Kang}, \binits{Y.}},
\bauthor{\bsnm{Kim}, \binits{J.}}:
\batitle{Chatmof: An artificial intelligence system for predicting and
  generating metal-organic frameworks using large language models}.
\bjtitle{Nat. Commun.}
\bvolume{15},
\bfpage{4705}
(\byear{2024})
\end{barticle}
\endbibitem

\bibitem[\protect\citeauthoryear{Jin et~al.}{2024}]{GeneGPT}
\begin{botherref}
\oauthor{\bsnm{Jin}, \binits{Q.}},
\oauthor{\bsnm{Yang}, \binits{Y.}},
\oauthor{\bsnm{Chen}, \binits{Q.}},
\oauthor{\bsnm{Lu}, \binits{Z.}}:
Genegpt: augmenting large language models with domain tools for improved access
  to biomedical information.
Bioinformatics
\textbf{40}(2)
(2024)
\end{botherref}
\endbibitem

\bibitem[\protect\citeauthoryear{Liu et~al.}{2025}]{PDE-Agent}
\begin{botherref}
\oauthor{\bsnm{Liu}, \binits{J.}},
\oauthor{\bsnm{Zhu}, \binits{R.}},
\oauthor{\bsnm{Xu}, \binits{J.}},
\oauthor{\bsnm{Ding}, \binits{K.}},
\oauthor{\bsnm{Zhang}, \binits{X.-Y.}},
\oauthor{\bsnm{Meng}, \binits{G.}},
\oauthor{\bsnm{Liu}, \binits{C.-L.}}:
PDE-Agent: A toolchain-augmented multi-agent framework for PDE solving.
Preprint at \url{https://arxiv.org/abs/2512.16214}
(2025)
\end{botherref}
\endbibitem

\bibitem[\protect\citeauthoryear{Huang
  et~al.}{2024}]{huang2024codecottacklingcodesyntax}
\begin{botherref}
\oauthor{\bsnm{Huang}, \binits{D.}},
\oauthor{\bsnm{Bu}, \binits{Q.}},
\oauthor{\bsnm{Qing}, \binits{Y.}},
\oauthor{\bsnm{Cui}, \binits{H.}}:
CodeCoT: Tackling Code Syntax Errors in CoT Reasoning for Code Generation.
Preprint at \url{https://arxiv.org/abs/2308.08784}
(2024)
\end{botherref}
\endbibitem

\bibitem[\protect\citeauthoryear{Romera{-}Paredes et~al.}{2024}]{FunSearch}
\begin{barticle}
\bauthor{\bsnm{Romera{-}Paredes}, \binits{B.}},
\bauthor{\bsnm{Barekatain}, \binits{M.}},
\bauthor{\bsnm{Novikov}, \binits{A.}},
\bauthor{\bsnm{Balog}, \binits{M.}},
\bauthor{\bsnm{Kumar}, \binits{M.P.}},
\bauthor{\bsnm{Dupont}, \binits{E.}},
\bauthor{\bsnm{Ruiz}, \binits{F.J.R.}},
\bauthor{\bsnm{Ellenberg}, \binits{J.S.}},
\bauthor{\bsnm{Wang}, \binits{P.}},
\bauthor{\bsnm{Fawzi}, \binits{O.}},
\bauthor{\bsnm{Kohli}, \binits{P.}},
\bauthor{\bsnm{Fawzi}, \binits{A.}}:
\batitle{Mathematical discoveries from program search with large language
  models}.
\bjtitle{Nat.}
\bvolume{625}(\bissue{7995}),
\bfpage{468}--\blpage{475}
(\byear{2024})
\end{barticle}
\endbibitem

\bibitem[\protect\citeauthoryear{Liu et~al.}{2024}]{EoH}
\begin{bchapter}
\bauthor{\bsnm{Liu}, \binits{F.}},
\bauthor{\bsnm{Tong}, \binits{X.}},
\bauthor{\bsnm{Yuan}, \binits{M.}},
\bauthor{\bsnm{Lin}, \binits{X.}},
\bauthor{\bsnm{Luo}, \binits{F.}},
\bauthor{\bsnm{Wang}, \binits{Z.}},
\bauthor{\bsnm{Lu}, \binits{Z.}},
\bauthor{\bsnm{Zhang}, \binits{Q.}}:
\bctitle{Evolution of heuristics: towards efficient automatic algorithm design
  using large language model}.
In: \bbtitle{Proceedings of the 41st International Conference on Machine
  Learning}
(\byear{2024})
\end{bchapter}
\endbibitem

\bibitem[\protect\citeauthoryear{Ding et~al.}{2025}]{EvoVLMA}
\begin{bchapter}
\bauthor{\bsnm{Ding}, \binits{K.}},
\bauthor{\bsnm{Wang}, \binits{Y.}},
\bauthor{\bsnm{Xiang}, \binits{S.}}:
\bctitle{Evovlma: Evolutionary vision-language model adaptation}.
In: \bbtitle{Proceedings of the 33rd ACM International Conference on
  Multimedia},
pp. \bfpage{4619}--\blpage{4628}
(\byear{2025})
\end{bchapter}
\endbibitem

\bibitem[\protect\citeauthoryear{Wang and Zeng}{2025}]{EvoMCTS}
\begin{botherref}
\oauthor{\bsnm{Wang}, \binits{H.}},
\oauthor{\bsnm{Zeng}, \binits{L.}}:
Automated Algorithmic Discovery for Scientific Computing through LLM-Guided
  Evolutionary Search: A Case Study in Gravitational-Wave Detection.
Preprint at \url{https://arxiv.org/abs/2508.03661}
(2025)
\end{botherref}
\endbibitem

\bibitem[\protect\citeauthoryear{Chen et~al.}{2024}]{ChenLSZ24}
\begin{bchapter}
\bauthor{\bsnm{Chen}, \binits{X.}},
\bauthor{\bsnm{Lin}, \binits{M.}},
\bauthor{\bsnm{Sch{\"{a}}rli}, \binits{N.}},
\bauthor{\bsnm{Zhou}, \binits{D.}}:
\bctitle{Teaching large language models to self-debug}.
In: \bbtitle{The Twelfth International Conference on Learning Representations}
(\byear{2024})
\end{bchapter}
\endbibitem

\bibitem[\protect\citeauthoryear{Shinn et~al.}{2023}]{ShinnCGNY23}
\begin{barticle}
\bauthor{\bsnm{Shinn}, \binits{N.}},
\bauthor{\bsnm{Cassano}, \binits{F.}},
\bauthor{\bsnm{Gopinath}, \binits{A.}},
\bauthor{\bsnm{Narasimhan}, \binits{K.}},
\bauthor{\bsnm{Yao}, \binits{S.}}:
\batitle{Reflexion: Language agents with verbal reinforcement learning}.
\bjtitle{Advances in Neural Information Processing Systems}
\bvolume{36},
\bfpage{8634}--\blpage{8652}
(\byear{2023})
\end{barticle}
\endbibitem

\bibitem[\protect\citeauthoryear{Islam et~al.}{2024}]{IslamAP24}
\begin{bchapter}
\bauthor{\bsnm{Islam}, \binits{M.A.}},
\bauthor{\bsnm{Ali}, \binits{M.E.}},
\bauthor{\bsnm{Parvez}, \binits{M.R.}}:
\bctitle{Mapcoder: Multi-agent code generation for competitive problem
  solving}.
In: \bbtitle{Proceedings of the 62nd Annual Meeting of the Association for
  Computational Linguistics},
pp. \bfpage{4912}--\blpage{4944}
(\byear{2024})
\end{bchapter}
\endbibitem

\bibitem[\protect\citeauthoryear{Stienstra
  et~al.}{2024}]{stienstra2024graphormer}
\begin{barticle}
\bauthor{\bsnm{Stienstra}, \binits{C.M.}},
\bauthor{\bsnm{Hebert}, \binits{L.}},
\bauthor{\bsnm{Thomas}, \binits{P.}},
\bauthor{\bsnm{Haack}, \binits{A.}},
\bauthor{\bsnm{Guo}, \binits{J.}},
\bauthor{\bsnm{Hopkins}, \binits{W.S.}}:
\batitle{Graphormer-ir: Graph transformers predict experimental ir spectra
  using highly specialized attention}.
\bjtitle{Journal of chemical information and modeling}
\bvolume{64}(\bissue{12}),
\bfpage{4613}--\blpage{4629}
(\byear{2024})
\end{barticle}
\endbibitem

\bibitem[\protect\citeauthoryear{Bhatia et~al.}{2025a}]{bhatia2025leveraging}
\begin{barticle}
\bauthor{\bsnm{Bhatia}, \binits{N.}},
\bauthor{\bsnm{Rinke}, \binits{P.}},
\bauthor{\bsnm{Krej{\v{c}}{\'\i}}, \binits{O.}}:
\batitle{Leveraging active learning-enhanced machine-learned interatomic
  potential for efficient infrared spectra prediction}.
\bjtitle{npj Computational Materials}
\bvolume{11}(\bissue{1}),
\bfpage{324}
(\byear{2025})
\end{barticle}
\endbibitem

\bibitem[\protect\citeauthoryear{Bhatia et~al.}{2025b}]{MACE4IR}
\begin{botherref}
\oauthor{\bsnm{Bhatia}, \binits{N.}},
\oauthor{\bsnm{Krejci}, \binits{O.}},
\oauthor{\bsnm{Botti}, \binits{S.}},
\oauthor{\bsnm{Rinke}, \binits{P.}},
\oauthor{\bsnm{Marques}, \binits{M.A.L.}}:
MACE4IR: A foundation model for molecular infrared spectroscopy.
Preprint at \url{https://arxiv.org/abs/2508.19118}
(2025)
\end{botherref}
\endbibitem

\bibitem[\protect\citeauthoryear{Ding et~al.}{2025}]{ding2025scitoolagent}
\begin{botherref}
\oauthor{\bsnm{Ding}, \binits{K.}},
\oauthor{\bsnm{Yu}, \binits{J.}},
\oauthor{\bsnm{Huang}, \binits{J.}},
\oauthor{\bsnm{Yang}, \binits{Y.}},
\oauthor{\bsnm{Zhang}, \binits{Q.}},
\oauthor{\bsnm{Chen}, \binits{H.}}:
Scitoolagent: a knowledge-graph-driven scientific agent for multitool
  integration.
Nat. Comput. Sci.
(2025)
\end{botherref}
\endbibitem

\bibitem[\protect\citeauthoryear{Mialon et~al.}{2023}]{GAIA}
\begin{botherref}
\oauthor{\bsnm{Mialon}, \binits{G.}},
\oauthor{\bsnm{Fourrier}, \binits{C.}},
\oauthor{\bsnm{Swift}, \binits{C.}},
\oauthor{\bsnm{Wolf}, \binits{T.}},
\oauthor{\bsnm{LeCun}, \binits{Y.}},
\oauthor{\bsnm{Scialom}, \binits{T.}}:
{GAIA}: a benchmark for General AI Assistants.
Preprint at \url{https://arxiv.org/abs/2311.12983}
(2023)
\end{botherref}
\endbibitem

\end{thebibliography}

\cleardoublepage

\begin{appendices}
	\section{Evaluation Method}\label{secA:Eval}
	Performance assessment relied on a suite of metrics derived from human-defined evaluation code. This code encodes task-specific success criteria explicitly. It validates both functional correctness, e.g. output format and file existence, and scientific plausibility, including physically reasonable value ranges and adherence to domain constraints. For assistant-type tasks, the evaluation function validates solution outputs and returns a binary score: 0 for incorrect results and 1 for correct ones. For optimization-type tasks, it computes a floating-point score to quantify solution quality. Higher values indicate better performance.
	
	Reported metrics capture different dimensions of solution quality. For assistant-type problems, we report three metrics. First, \textbf{Pass@1} accuracy measures syntactic executability by checking if generated code runs without errors. Second, \textbf{Tool Acc.} evaluates tool selection accuracy. Third, \textbf{Accuracy} serves as the primary metric. It is calculated by running human-defined evaluation code on solution outputs, distinguishing merely executable code from functionally correct solutions.
	
	For optimization-type problems, each task is split into training and test subsets. This design evaluates both optimization efficacy and generalization ability. We report six metrics: \textbf{Pass@1}, \textbf{Tool Acc.}, \textbf{Train Score}, \textbf{Test Score}, \textbf{Train Rank}, and \textbf{Test Rank}. Train Score and Test Score are averages of scores from human-defined evaluation code executed on the respective subsets. Train Score quantifies how well the solution optimizes the objective using given data. Test Score specifically measures generalization to unseen inputs within the same problem class, assessing robustness. Train Rank and Test Rank are normalized ranking scores. They mitigate scale discrepancies across different tasks, making them the primary metrics for evaluating optimization performance.
	
	Aggregated metrics, \textbf{Overall Pass@1}, \textbf{Overall Score}, and \textbf{Overall Tool Acc.}, provide a holistic summary across all tasks. Overall Score is a weighted mean of Accuracy, Train Rank and Test Rank. Detailed formulations of all metrics are provided below. We conducted three independent trials per task. Results are reported as means with corresponding standard deviations.
	
	\bmhead{Notation}Let $N_a$ and $N_o$ denote the number of assistant-type and optimization-type tasks respectively. For task $i$, $T_i^{\text{pred}}$ represents the set of tools predicted by the solution code, $T_i^{\text{ref}}$ the reference tool set, $\mathbf{o}_i$ the generated outputs, and $\mathbf{o}_i^{\text{ref}}$ the reference outputs. For optimization-type tasks with training-test splits, $M_j^{\text{train}}$ and $M_j^{\text{test}}$ denote the number of training and test instances for problem $j$, with $\mathbf{o}_{jk}^{\text{train}}$ and $\mathbf{y}_{jk}^{\text{train}}$ representing outputs and targets for training instance $k$, and $\mathbf{o}_{jl}^{\text{test}}$ and $\mathbf{y}_{jl}^{\text{test}}$ for test instance $l$. Evaluation functions $f_i^{a}$ and $f_j^{o}$ compute task-specific quality scores, normalized around 1.
	
	\bmhead{Assistant Task Metrics}For assistant-type problems, three metrics are reported. Pass@1 accuracy ($A_{\text{pass}}^{a}$) validates solution executability, where $\mathbb{I}_{\{\text{execution\_success}\}}(i)=1$ if the solution code executes without critical errors and 0 otherwise:
	\begin{equation}
		A_{\text{pass}}^{a} = \frac{1}{N_a} \sum_{i=1}^{N_a} \mathbb{I}_{\{\text{execution\_success}\}}(i).
	\end{equation}
	
	Tool selection accuracy ($A_{\text{tool}}^{a}$) measures tool identification accuracy through intersection-over-prediction:
	\begin{equation}
		A_{\text{tool}}^{a} = \frac{1}{N_a} \sum_{i=1}^{N_a} \frac{|T_i^{\text{pred}} \cap T_i^{\text{ref}}|}{|T_i^{\text{pred}}|}.
	\end{equation}
	
	Solution output accuracy ($A_{\text{accuracy}}^{a}$) quantifies output correctness through task-specific evaluation:
	\begin{equation}
		A_{\text{accuracy}}^{a} = \frac{1}{N_a} \sum_{i=1}^{N_a} f_i^{a}(\mathbf{o}_i, \mathbf{o}_i^{\text{ref}}).
	\end{equation}
	
	\textbf{Optimization Task Metrics:} For performance optimization problems with training-test splits, six metrics are reported. Pass@1 accuracy ($A_{\text{pass}}^{o}$) assesses solution executability:
	\begin{equation}
		A_{\text{pass}}^{o} = \frac{1}{N_o} \sum_{j=1}^{N_o} \mathbb{I}_{\{\text{execution\_success}\}}(j).
	\end{equation}
	
	Tool selection accuracy ($A_{\text{tool}}^{o}$) evaluates tool identification:
	\begin{equation}
		A_{\text{tool}}^{o} = \frac{1}{N_o} \sum_{j=1}^{N_o} \frac{|T_j^{\text{pred}} \cap T_j^{\text{ref}}|}{|T_j^{\text{pred}}|}.
	\end{equation}
	
	Training score ($S_{\text{train}}^{o}$) measures optimization performance on training data:
	\begin{equation}
		S_{\text{train}}^{o} = \frac{1}{N_o} \sum_{j=1}^{N_o} \frac{1}{M_j^{\text{train}}} \sum_{k=1}^{M_j^{\text{train}}} f_j^{o}(\mathbf{o}_{jk}^{\text{train}}, \mathbf{y}_{jk}^{\text{train}}).
	\end{equation}
	
	Test accuracy ($S_{\text{test}}^{o}$) quantifies generalization capability on unseen instances:
	\begin{equation}
		S_{\text{test}}^{o} = \frac{1}{N_o} \sum_{j=1}^{N_o} \frac{1}{M_j^{\text{test}}} \sum_{l=1}^{M_j^{\text{test}}} f_j^{o}(\mathbf{o}_{jl}^{\text{test}}, \mathbf{y}_{jl}^{\text{test}}).
	\end{equation}
	
	Besides, we also compute the averaged ranking score of the training score on each case:
	\begin{equation}
		R_{\text{train}}^{o} = \frac{1}{N_o} \sum_{j=1}^{N_o} \text{Rank}\Big(\frac{1}{M_j^{\text{train}}} \sum_{k=1}^{M_j^{\text{train}}} f_j^{o}(\mathbf{o}_{jk}^{\text{train}}, \mathbf{y}_{jk}^{\text{train}})\Big),
	\end{equation}
	where $\text{Rank}()$ denote the function computes the normalized rank, which is defined as:
	\begin{equation}
		\text{Rank}(s) = \frac{K - r(s)}{K - 1}.
	\end{equation}
	Here $r(s)$ is the rank of score $s$ among all $K$ compared methods, with rank 1 assigned to the best (highest) score and rank $K$ to the worst (lowest) score. When multiple methods achieve identical scores, they are assigned the average rank of their tied positions. The normalized rank $\text{Rank}(s)$ then maps the raw rank $r(s)$ to the range $[0, 1]$, where 1 represents the best performance and 0 represents the worst performance among all $K$ methods.
	
	The averaged ranking score of the test score is defined similarly:
	\begin{equation}
		R_{\text{test}}^{o} = \frac{1}{N_o} \sum_{j=1}^{N_o} \frac{1}{M_j^{\text{test}}}\text{Rank}\Big(\sum_{l=1}^{M_j^{\text{test}}} f_j^{o}(\mathbf{o}_{jl}^{\text{test}}. \mathbf{y}_{jl}^{\text{test}})\Big).
	\end{equation}
	
	\bmhead{Composite Performance Scores}The evaluation framework employs distinct metric compositions for assistant and optimization-type tasks while providing cross-task averages for comparability. The overall solution quality score ($S_{\text{quality}}$) integrates performance across task types:
	\begin{equation}
		S_{\text{quality}} = 0.5 \cdot A_{\text{accuracy}}^{a} + 0.25 \cdot R_{\text{train}}^{o} + 0.25 \cdot R_{\text{test}}^{o},
	\end{equation}
	where $A_{\text{accuracy}}^{a}$ represents assistant-type task output accuracy, $R_{\text{train}}^{o}$ denotes the normalized ranking score of the training score, and $R_{\text{test}}^{o}$ indicates the normalized ranking score of the test score.
	
	The Overall Pass@1 rate across two task types is defined as:
	\begin{equation}
		A_{\text{pass}} = 0.5 \cdot A_{\text{pass}}^{a} + 0.5 \cdot A_{\text{pass}}^{o},
	\end{equation}
	where $A_{\text{pass}}^{a}$ and $A_{\text{pass}}^{o}$ denote the Pass@1 rate on assistant-type and optimization-type tasks, respectively.
	
	The overall tool selection performance is evaluated by:
	\begin{equation}
		A_{\text{tool}} = 0.5 \cdot A_{\text{tool}}^{a} + 0.5 \cdot A_{\text{tool}}^{o},
	\end{equation}
	where $A_{\text{tool}}^{a} $ and $A_{\text{tool}}^{o}$ denote the tool selection accuracy on assistant-type and optimization-type tasks, respectively.
	
	\bmhead{Prompt Template for Parameter Parsing} During evaluation of optimization-type tasks, proper parameter initialization for the \texttt{solve} function is essential for executing solution code on test subset. Manual parameter specification proves impractical at scale, prompting the adoption of an automated parsing approach using DeepSeek-V3.2. This method employs the structured prompt shown below, which accepts test question descriptions and solution code as inputs, then extracts a JSON-formatted keyword parameter dictionary (\texttt{kwargs}). The parsed parameters enable automated test execution via \texttt{solve(tools, **kwargs)}, facilitating systematic evaluation across diverse optimization problems without manual intervention.
	
	\begin{prompt}[Prompt for parameter parsing.]{prompt:param_parse}
		Here is a query question and a python code solving it. The optional arguments of this code should be determined based on the question. Help me extract the needed optional arguments.
		
		The query problem: {question}
		
		The python solution code: {solution_code}
		
		Return a python dictionary containing your extracted optional parameters. The keys are the optional parameter names in the code, the values should be set according to the question.
		Your result should be enclosed between ```python and ```. For example (do not directly copy):
		
		```python
		{"arg_name1": arg1_value, "arg_name2": arg2_value}
		```
		
		Notice:
		* Do not give extra explaination
		* There should be only one line between ```python and ```
		* Keys should match the optional argument names
		* Do not consider the positional arguments
		* If some optional argument values cannot be determined by the question directly, do not include them. For example, if the image_size parameter is not mentioned, do not set it by yourself.
	\end{prompt}
	
	\section{More Results}\label{secA:Results}
	
	\bmhead{Case 2} The solution code of case 2  is given in Code~\ref{code:case2_code} and the generated evaluation code is given in Code~\ref{code:case2_eval}.
	\begin{code}[Case 2's solution code.]{code:case2_code}
		# import Python packages
		import numpy as np
		import pandas as pd
		import matplotlib.pyplot as plt
		
		# define helper function1: load and process IR spectrum data
		def load_and_process_spectrum(file_path, source_type="lammps"):
		"""
		Load IR spectrum data from file and process it.
		
		Args:
		file_path (str): Path to the spectrum file
		source_type (str): Type of spectrum source - "lammps" or "ml"
		
		Returns:
		tuple: (frequencies, intensities) arrays
		"""
		if source_type == "lammps":
		# Load LAMMPS spectrum CSV
		df = pd.read_csv(file_path)
		frequencies = df['Frequency(cm^-1)'].values
		intensities = df['Spectra'].values
		else:  # ml prediction
		# Load ML predicted spectrum CSV
		df = pd.read_csv(file_path)
		frequencies = df['Frequency'].values
		intensities = df['Intensity'].values
		
		return frequencies, intensities
		
		# define main solve function
		def solve(tools, smiles="CCO", xyz_file_path="tasks/task120/mol.xyz", 
		charge=0.0, multiplicity=1.0, steps=500000, box_dim=15.0, 
		temperature=300.0, data_points=8339, average=1):
		"""
		Compare IR spectrum from molecular dynamics simulation with ML prediction.
		
		Args:
		- tools (dict): Dictionary containing available tools
		- smiles (str): SMILES string of the molecule (default: "CCO")
		- xyz_file_path (str): Path to the 3D molecular structure file (default: "tasks/task120/mol.xyz")
		- charge (float): Charge of the molecule (default: 0.0)
		- multiplicity (float): Multiplicity of the molecule (default: 1.0)
		- steps (int): Number of MD simulation steps (default: 500000)
		- box_dim (float): Size of cubic simulation box in Angstroms (default: 15.0)
		- temperature (float): Temperature in Kelvin for IR calculation (default: 300.0)
		- data_points (int): Number of data points for IR analysis (default: 8339)
		- average (int): Whether to use average parameters for ML prediction (default: 1)
		
		Returns:
		- str: Path to the saved visualization image comparing both IR spectra
		"""
		
		# Step 1: Generate LAMMPS simulation input files from SMILES string using EMC setup
		emc_result = tools["EMC_Setup_Tool"].execute(smiles=smiles, field="pcff")
		setup_data_path = emc_result["setup_data_path"]
		setup_params_path = emc_result["setup_params_path"]
		
		# Step 2: Run LAMMPS molecular dynamics simulation to generate dipole moment time-series data
		lammps_result = tools["LAMMPS_Simulation_Tool"].execute(
		setup_data_path=setup_data_path,
		setup_params_path=setup_params_path,
		steps=steps,
		box_dim=box_dim
		)
		dipole_path = lammps_result["dipole_path"]
		
		# Step 3: Calculate IR spectrum from LAMMPS dipole moment data
		lammps_ir_result = tools["LAMMPS_Dipole_IRSpectra_Tool"].execute(
		dipole_file_path=dipole_path,
		temperature=temperature,
		data_points=data_points
		)
		lammps_spectrum_path = lammps_ir_result["ir_spectra_file"]
		
		# Step 4: Predict IR spectrum from 3D molecular structure using machine learning models
		ml_ir_result = tools["IRspectrum_Predictor_Use_3D_Structure_Tool"].execute(
		input_file_name=xyz_file_path,
		charge=charge,
		multiplicity=multiplicity,
		average=average
		)
		ml_spectrum_path = ml_ir_result["ir_spectrum_path"]
		
		# Step 5: Load and process both IR spectra
		# Load LAMMPS spectrum
		lammps_freq, lammps_int = load_and_process_spectrum(lammps_spectrum_path, "lammps")
		
		# Load ML predicted spectrum
		ml_freq, ml_int = load_and_process_spectrum(ml_spectrum_path, "ml")
		
		# Filter to 400-4000 cm^-1 range
		lammps_mask = (lammps_freq >= 400) & (lammps_freq <= 4000)
		ml_mask = (ml_freq >= 400) & (ml_freq <= 4000)
		
		lammps_freq_filtered = lammps_freq[lammps_mask]
		lammps_int_filtered = lammps_int[lammps_mask]
		
		ml_freq_filtered = ml_freq[ml_mask]
		ml_int_filtered = ml_int[ml_mask]
		
		# Normalize to maximum value of 1.0
		if len(lammps_int_filtered) > 0:
		lammps_int_normalized = lammps_int_filtered / np.max(lammps_int_filtered)
		else:
		lammps_int_normalized = np.array([])
		
		if len(ml_int_filtered) > 0:
		ml_int_normalized = ml_int_filtered / np.max(ml_int_filtered)
		else:
		ml_int_normalized = np.array([])
		
		# Step 6: Create comparative visualization
		plt.figure(figsize=(10, 6))
		
		if len(lammps_freq_filtered) > 0 and len(lammps_int_normalized) > 0:
		plt.plot(lammps_freq_filtered, lammps_int_normalized, label='MD Simulation', linewidth=2, alpha=0.8)
		
		if len(ml_freq_filtered) > 0 and len(ml_int_normalized) > 0:
		plt.plot(ml_freq_filtered, ml_int_normalized, label='ML Prediction', linewidth=2, alpha=0.8, linestyle='--')
		
		plt.xlabel('Wavenumber (cm$^{-1}$)', fontsize=12)
		plt.ylabel('Normalized Intensity', fontsize=12)
		plt.title('Comparison of IR Spectra: MD Simulation vs ML Prediction', fontsize=14)
		plt.legend(fontsize=11)
		plt.grid(True, alpha=0.3)
		plt.xlim(400, 4000)
		plt.ylim(0, 1.1)
		
		# Save the visualization
		output_image_path = "ir_spectra_comparison.png"
		plt.tight_layout()
		plt.savefig(output_image_path, dpi=300, bbox_inches='tight')
		plt.close()
		
		return output_image_path
	\end{code}
	
	\begin{code}[Case 2's generated evaluation code.]{code:case2_eval}
		def evaluate(result):
		"""
		Evaluate whether the solution result meets the problem requirements.
		
		Args:
		- result (str): Path to the visualization image comparing IR spectra
		
		Returns:
		- int: Return 1 if result is correct; otherwise, an error will be raised
		"""
		# IMPORTS
		import os
		import cv2
		import numpy as np
		import pandas as pd
		
		# DATA PREPARATION
		# Check if result is a valid file path
		assert isinstance(result, str), f"Result must be a string file path, got {type(result)}"
		assert os.path.exists(result), f"Image file does not exist: {result}"
		
		# Read the image
		img = cv2.imread(result)
		assert img is not None, f"Failed to read image from path: {result}"
		
		# VALIDATION LOGIC
		# 1. Validate the returned image file
		assert os.path.isfile(result), f"Result must be a file path, not a directory: {result}"
		assert result.lower().endswith(('.png', '.jpg', '.jpeg', '.tiff', '.bmp')), \
		f"Result must be an image file, got: {result}"
		
		# 2. Validate image content and properties
		assert img.shape[0] > 0 and img.shape[1] > 0, "Image must have non-zero dimensions"
		assert len(img.shape) == 3, "Image must be a 3-channel color image"
		assert img.shape[2] == 3, "Image must have 3 color channels (BGR)"
		
		# 3. Check that the image is not completely empty/black
		img_mean = np.mean(img)
		assert img_mean > 10, "Image appears to be mostly empty or black"
		
		# 4. Check that input structure file exists
		assert os.path.exists("tasks/task120/mol.xyz"), \
		"Input molecular structure file not found: tasks/task120/mol.xyz"
		
		# 5. Validate the image filename matches expected pattern
		# The solution returns "ir_spectra_comparison.png" by default
		# But we accept any valid image filename
		print(f"Successfully validated IR spectra comparison image: {result}")
		
		return 1
	\end{code}
	
	\bmhead{Case 3} The best solution codes at iteration 0, 3, 7 are given in Code~\ref{code:case3_i0}, Code~\ref{code:case3_i3} and Code~\ref{code:case3_i7}, respectively.
	
	\begin{code}[Case 3's solution code at iteration 0.]{code:case3_i0}
		import numpy as np
		
		# define main solve function
		def solve(
				tools,
				spectra_path="tasks/task231/spectra_train.npy",
				wavenumber_path="tasks/task231/wavenumber.npy",
				formulas_path="tasks/task231/formulas_train.txt",
				beam_size=10,
				n_best=10,
				min_length=5
			):
			"""
			Predict SMILES from IR spectra and molecular formulas using a trained NMT model.
			
			Args:
			- tools (dict): Dictionary containing available tools
			- spectra_path (str): Path to the spectra npy file (rows=molecules, cols=wavenumbers)
			- wavenumber_path (str): Path to the wavenumber npy file (1-d array)
			- formulas_path (str): Path to the formulas txt file (one formula per line)
			- beam_size (int): Beam size for beam search (default: 10)
			- n_best (int): Number of best translations to output (default: 10)
			- min_length (int): Minimum length of translation (default: 5)
			
			Returns:
			- list[str]: Paths to the predicted SMILES txt files
			"""
		
			# Step 1: Load the spectrum data, wavenumber data, and formulas data from the provided files.
			# Tools needed: []. Packages needed: [numpy].
			spectra = np.load(spectra_path)  # Shape: (n_molecules, n_wavenumbers)
			wavenumbers = np.load(wavenumber_path)  # Shape: (n_wavenumbers,)
			with open(formulas_path, 'r') as f:
				formulas = [line.strip() for line in f.readlines()]
			
			# Step 2: Process the loaded spectrum data and formulas data into the required format for the NMT_Formula_IRSpectrum_To_Smiles_Tool tool.
			# Tools needed: []. Packages needed: [numpy].
			# Create spectrum files (n*2 arrays: wavenumber + spectrum value)
			spectrum_files = []
			for i in range(len(formulas)):
				spectrum_data = np.column_stack((wavenumbers, spectra[i]))  # Combine wavenumbers and spectrum values
				spectrum_file = f"spectrum_{i}.npy"
				np.save(spectrum_file, spectrum_data)
				spectrum_files.append(spectrum_file)
			
			# Step 3: Use the NMT_Formula_IRSpectrum_To_Smiles_Tool tool to predict SMILES from the formatted spectrum data and formulas.
			# Tools needed: [NMT_Formula_IRSpectrum_To_Smiles_Tool]. Packages needed: [].
			execution = tools["NMT_Formula_IRSpectrum_To_Smiles_Tool"].execute(
				formulas=formulas,
				spectrum_files=spectrum_files,
				beam_size=beam_size,
				n_best=n_best,
				min_length=min_length
			)
			
			return execution["output_files"]
	\end{code}
	
	\begin{code}[Case 3's solution code at iteration 3.]{code:case3_i3}
		import numpy as np
		
		# define main solve function
		def solve(
				tools,
				spectra_path="tasks/task231/spectra_train.npy",
				wavenumber_path="tasks/task231/wavenumber.npy",
				formulas_path="tasks/task231/formulas_train.txt",
				beam_size=10,
				n_best=10,
				min_length=5
			):
			"""
			Predict SMILES from IR spectra using a trained NMT model that only requires spectrum (no formula).
			
			Args:
			- tools (dict): Dictionary containing available tools
			- spectra_path (str): Path to the spectra npy file (rows=molecules, cols=wavenumbers)
			- wavenumber_path (str): Path to the wavenumber npy file (1-d array)
			- formulas_path (str): Path to the formulas txt file (one formula per line) - used here just for counting number of molecules
			- beam_size (int): Beam size for beam search (default: 10)
			- n_best (int): Number of best translations to output (default: 10)
			- min_length (int): Minimum length of translation (default: 5)
			
			Returns:
			- list[str]: Paths to the predicted SMILES txt files
			"""
		
			# Step 1: Load the spectrum data and wavenumber data from the provided files.
			# Tools needed: []. Packages needed: [numpy].
			spectra = np.load(spectra_path)  # Shape: (n_molecules, n_wavenumbers)
			wavenumbers = np.load(wavenumber_path)  # Shape: (n_wavenumbers,)
			
			# We don't actually need the formulas themselves, but we do need the count of molecules
			with open(formulas_path, 'r') as f:
				formulas = [line.strip() for line in f.readlines()]
			
			# Ensure the number of spectra matches the number of formulas
			assert len(spectra) == len(formulas), "Mismatch between number of spectra and formulas"
			
			# Step 2: Process the loaded spectrum data into the required format for the NMT_IRSpectrum_To_Smiles_Tool tool.
			# Tools needed: []. Packages needed: [numpy].
			# Create spectrum files (n*2 arrays: wavenumber + spectrum value)
			spectrum_files = []
			for i in range(len(spectra)):
				spectrum_data = np.column_stack((wavenumbers, spectra[i]))  # Combine wavenumbers and spectrum values
				spectrum_file = f"spectrum_{i}.npy"
				np.save(spectrum_file, spectrum_data)
				spectrum_files.append(spectrum_file)
			
			# Step 3: Use the NMT_IRSpectrum_To_Smiles_Tool tool to predict SMILES from the formatted spectrum data.
			# This tool does not require formulas.
			# Tools needed: [NMT_IRSpectrum_To_Smiles_Tool]. Packages needed: [].
			execution = tools["NMT_IRSpectrum_To_Smiles_Tool"].execute(
				spectrum_files=spectrum_files,
				beam_size=beam_size,
				n_best=n_best,
				min_length=min_length
			)
			
			return execution["output_files"]
	\end{code}

	\begin{code}[Case 3's solution code at iteration 7.]{code:case3_i7}
		import numpy as np
		
		# define main solve function
		def solve(
				tools,
				spectra_path="tasks/task231/spectra_train.npy",
				wavenumber_path="tasks/task231/wavenumber.npy",
				formulas_path="tasks/task231/formulas_train.txt",
				beam_size=20,
				n_best=20,
				min_length=10
			):
			"""
			Predict SMILES from IR spectra using a trained NMT model that only requires spectrum (no formula).
			
			Args:
			- tools (dict): Dictionary containing available tools
			- spectra_path (str): Path to the spectra npy file (rows=molecules, cols=wavenumbers)
			- wavenumber_path (str): Path to the wavenumber npy file (1-d array)
			- formulas_path (str): Path to the formulas txt file (one formula per line) - used here just for counting number of molecules
			- beam_size (int): Increased beam size for beam search to explore more candidates (default: 20)
			- n_best (int): Increased number of best translations to output for better diversity (default: 20)
			- min_length (int): Increased minimum length of translation to avoid short predictions (default: 10)
			
			Returns:
			- list[str]: Paths to the predicted SMILES txt files
			"""
			
			# Step 1: Load the spectrum data and wavenumber data from the provided files.
			# Tools needed: []. Packages needed: [numpy].
			spectra = np.load(spectra_path)  # Shape: (n_molecules, n_wavenumbers)
			wavenumbers = np.load(wavenumber_path)  # Shape: (n_wavenumbers,)
			
			# We don't actually need the formulas themselves, but we do need the count of molecules
			with open(formulas_path, 'r') as f:
				formulas = [line.strip() for line in f.readlines()]
			
			# Ensure the number of spectra matches the number of formulas
			assert len(spectra) == len(formulas), "Mismatch between number of spectra and formulas"
			
			# Step 2: Process the loaded spectrum data into the required format for the NMT_IRSpectrum_To_Smiles_Tool tool.
			# Tools needed: []. Packages needed: [numpy].
			# Create spectrum files (n*2 arrays: wavenumber + spectrum value)
			spectrum_files = []
			for i in range(len(spectra)):
				spectrum_data = np.column_stack((wavenumbers, spectra[i]))  # Combine wavenumbers and spectrum values
				spectrum_file = f"spectrum_{i}.npy"
				np.save(spectrum_file, spectrum_data)
				spectrum_files.append(spectrum_file)
			
			# Step 3: Use the NMT_IRSpectrum_To_Smiles_Tool tool to predict SMILES from the formatted spectrum data.
			# This tool does not require formulas.
			# Tools needed: [NMT_IRSpectrum_To_Smiles_Tool]. Packages needed: [].
			execution = tools["NMT_IRSpectrum_To_Smiles_Tool"].execute(
				spectrum_files=spectrum_files,
				beam_size=beam_size,
				n_best=n_best,
				min_length=min_length
			)
			
			return execution["output_files"]
	\end{code}
	
	\bmhead{Case 4}The best solution codes at iteration 0, 2, 3, 9 are given in Code~\ref{code:case4_i0}, Code~\ref{code:case4_i2}, Code~\ref{code:case4_i3}, Code~\ref{code:case4_i9}, respectively.
	
	\begin{code}[Case 4's solution code at iteration 0.]{code:case4_i0}
		# import Python packages
		import os
		
		# define main solve function
		def solve(tools, rgb_image_paths: list = ["tasks/task212/vis_im1.jpg","tasks/task212/vis_im2.jpg", "tasks/task212/vis_im3.jpg","tasks/task212/vis_im4.jpg","tasks/task212/vis_im5.jpg"], seed: int = 1234):
		"""
		Generate fake thermal images from RGB images using PearlGAN style transfer.
		
		Args:
		- tools (dict): Dictionary containing available tools
		- rgb_image_paths (list[str]): Paths to the input RGB image files
		- seed (int): Seed for random number generator
		
		Returns:
		- list[str]: Paths to the generated fake thermal images
		"""
		
		# Step 1: Generate fake thermal images from the five RGB images using an RGB-to-thermal translation tool
		generated_paths = []
		
		for rgb_path in rgb_image_paths:
		# Use PearlGAN tool to generate fake thermal image
		execution = tools["PearlGAN_FLIR_RGB2T_Generation_Tool"].execute(
		input_image_path=rgb_path,
		seed=seed
		)
		generated_paths.append(execution["output_image_path"])
		
		# Step 2: Collect and return the paths of all generated thermal images as a list
		return generated_paths
	\end{code}
	
	\begin{code}[Case 4's solution code at iteration 2.]{code:case4_i2}
		# import Python packages
		import os
		
		# define main solve function
		def solve(tools, rgb_image_paths: list = ["tasks/task212/vis_im1.jpg", "tasks/task212/vis_im2.jpg", "tasks/task212/vis_im3.jpg", "tasks/task212/vis_im4.jpg", "tasks/task212/vis_im5.jpg"], seed: int = 1234):
		"""
		Generate fake thermal images from RGB images using FoalGAN style transfer.
		
		This function translates RGB images to thermal-style images using the FoalGAN model
		trained on FLIR dataset, which is specifically designed for thermal image generation
		with feedback-based object appearance learning.
		
		Args:
		- tools (dict): Dictionary containing available tools
		- rgb_image_paths (list[str]): Paths to the input RGB image files
		- seed (int): Seed for random number generator for reproducible results
		
		Returns:
		- list[str]: Paths to the generated fake thermal images in PNG format
		"""
		
		# Validate that the required tool is available
		if "FoalGAN_FLIR_RGB2T_Generation_Tool" not in tools:
		raise ValueError("Required tool 'FoalGAN_FLIR_RGB2T_Generation_Tool' not available")
		
		# Initialize list to store generated thermal image paths
		generated_thermal_paths = []
		
		# Process each RGB image to generate corresponding thermal image
		for rgb_path in rgb_image_paths:
		# Check if input file exists
		if not os.path.exists(rgb_path):
		raise FileNotFoundError(f"RGB image not found: {rgb_path}")
		
		# Apply FoalGAN style transfer to convert RGB to thermal
		execution_result = tools["FoalGAN_FLIR_RGB2T_Generation_Tool"].execute(
		input_image_path=rgb_path,
		seed=seed
		)
		
		# Extract the output path from the execution result
		thermal_output_path = execution_result["output_image_path"]
		generated_thermal_paths.append(thermal_output_path)
		
		return generated_thermal_paths
	\end{code}
	
	\begin{code}[Case 4's solution code at iteration 3.]{code:case4_i3}
		# import Python packages
		import os
		
		# define main solve function
		def solve(tools, rgb_image_paths: list = ["tasks/task212/vis_im1.jpg","tasks/task212/vis_im2.jpg", "tasks/task212/vis_im3.jpg","tasks/task212/vis_im4.jpg","tasks/task212/vis_im5.jpg"], steps: int = 500, seed: int = 1234):
		"""
		Generate fake thermal images from RGB images using Physics-Informed Diffusion model.
		
		This function translates RGB images to thermal-style images using the PID model
		trained on FLIR dataset, which incorporates physical constraints for more accurate
		thermal image generation compared to purely data-driven approaches.
		
		Args:
		- tools (dict): Dictionary containing available tools
		- rgb_image_paths (list[str]): Paths to the input RGB image files
		- steps (int): Number of steps for the diffusion process (higher = better quality)
		- seed (int): Seed for random number generator for reproducible results
		
		Returns:
		- list[str]: Paths to the generated fake thermal images in PNG format
		"""
		
		# Validate that the required tool is available
		if "PID_FLIR_RGB2T_Generation_Tool" not in tools:
		raise ValueError("Required tool 'PID_FLIR_RGB2T_Generation_Tool' not available")
		
		# Initialize list to store generated thermal image paths
		generated_thermal_paths = []
		
		# Process each RGB image to generate corresponding thermal image
		for rgb_path in rgb_image_paths:
		# Check if input file exists
		if not os.path.exists(rgb_path):
		raise FileNotFoundError(f"RGB image not found: {rgb_path}")
		
		# Apply PID diffusion model to convert RGB to thermal
		execution_result = tools["PID_FLIR_RGB2T_Generation_Tool"].execute(
		input_image_path=rgb_path,
		steps=steps,
		seed=seed
		)
		
		# Extract the output path from the execution result
		thermal_output_path = execution_result["output_image_path"]
		generated_thermal_paths.append(thermal_output_path)
		
		return generated_thermal_paths
	\end{code}
	
	\begin{code}[Case 4's solution code at iteration 9.]{code:case4_i9}
		# import Python packages
		import os
		import cv2
		import numpy as np
		
		# define helper function: calculate image complexity based on edge density
		def calculate_image_complexity(image_path):
		"""
		Calculate image complexity based on edge density using Canny edge detection.
		
		Args:
		image_path (str): Path to the input image
		
		Returns:
		float: Complexity score between 0 and 1
		"""
		try:
		# Read image in grayscale
		img = cv2.imread(image_path, cv2.IMREAD_GRAYSCALE)
		if img is None:
		return 0.5  # Default medium complexity if image cannot be read
		
		# Calculate edge density using Canny edge detection
		edges = cv2.Canny(img, 100, 200)
		edge_density = np.sum(edges > 0) / (img.shape[0] * img.shape[1])
		
		# Normalize to 0-1 range
		complexity = min(1.0, max(0.1, edge_density * 10))
		return complexity
		except Exception:
		return 0.5  # Default on error
		
		# define main solve function
		def solve(tools, rgb_image_paths: list = ["tasks/task212/vis_im1.jpg","tasks/task212/vis_im2.jpg", "tasks/task212/vis_im3.jpg","tasks/task212/vis_im4.jpg","tasks/task212/vis_im5.jpg"], seed: int = 1234, base_steps: int = 200):
		"""
		Generate fake thermal images from RGB images using Physics-Informed Diffusion model.
		
		This function translates RGB images to thermal-style images using the PID model
		trained on FLIR dataset, which incorporates physical constraints for more accurate
		thermal image generation. Image complexity is used to adapt diffusion steps.
		
		Args:
		- tools (dict): Dictionary containing available tools
		- rgb_image_paths (list[str]): Paths to the input RGB image files
		- seed (int): Seed for random number generator for reproducible results
		- base_steps (int): Base number of steps for diffusion process
		
		Returns:
		- list[str]: Paths to the generated fake thermal images in PNG format
		"""
		
		# Initialize list to store generated thermal image paths
		generated_thermal_paths = []
		
		# Check for available tools in order of preference
		preferred_tool = "PID_FLIR_RGB2T_Generation_Tool"
		fallback_tool = "FoalGAN_FLIR_RGB2T_Generation_Tool"
		
		# Determine which tool to use
		if preferred_tool in tools:
		selected_tool = preferred_tool
		elif fallback_tool in tools:
		selected_tool = fallback_tool
		else:
		raise ValueError(f"Neither {preferred_tool} nor {fallback_tool} available in tools")
		
		# Process each RGB image to generate corresponding thermal image
		for rgb_path in rgb_image_paths:
		# Check if input file exists
		if not os.path.exists(rgb_path):
		raise FileNotFoundError(f"RGB image not found: {rgb_path}")
		
		# Calculate image complexity to adapt diffusion steps
		complexity = calculate_image_complexity(rgb_path)
		
		# Adjust steps based on complexity (more complex images need more steps)
		adaptive_steps = int(base_steps * (0.8 + 0.4 * complexity))
		
		try:
		if selected_tool == "PID_FLIR_RGB2T_Generation_Tool":
		# Apply PID diffusion model for RGB-to-thermal translation
		execution_result = tools[selected_tool].execute(
		input_image_path=rgb_path,
		steps=adaptive_steps,
		seed=seed
		)
		else:
		# Apply fallback GAN-based tool
		execution_result = tools[selected_tool].execute(
		input_image_path=rgb_path,
		seed=seed
		)
		
		# Extract the output path from the execution result
		thermal_output_path = execution_result["output_image_path"]
		
		# Verify output file was created and has content
		if os.path.exists(thermal_output_path) and os.path.getsize(thermal_output_path) > 0:
		generated_thermal_paths.append(thermal_output_path)
		else:
		raise RuntimeError(f"Generated thermal image is empty or not created: {thermal_output_path}")
		
		except Exception as e:
		# If primary tool fails and we haven't tried fallback, try fallback
		if selected_tool == preferred_tool and fallback_tool in tools:
		try:
		execution_result = tools[fallback_tool].execute(
		input_image_path=rgb_path,
		seed=seed
		)
		thermal_output_path = execution_result["output_image_path"]
		if os.path.exists(thermal_output_path) and os.path.getsize(thermal_output_path) > 0:
		generated_thermal_paths.append(thermal_output_path)
		else:
		raise RuntimeError(f"Fallback tool also failed for: {rgb_path}")
		except Exception as fallback_error:
		raise RuntimeError(f"Both primary and fallback tools failed for {rgb_path}: {fallback_error}")
		else:
		raise RuntimeError(f"Tool execution failed for {rgb_path}: {e}")
		
		return generated_thermal_paths
	\end{code}
\end{appendices}

\end{document}